\documentclass[runningheads]{llncs}
\usepackage[svgnames]{xcolor}
\usepackage{graphicx}
\usepackage{lineno}
\usepackage{multirow}
\usepackage{tikz}
\usepackage{comment}
\usepackage{amsmath,amssymb} % define this before the line numbering.
\usepackage{bbm}
\usepackage[hyperfootnotes=false]{hyperref}
\usepackage{engord}
\usepackage{booktabs}
\usepackage{MnSymbol}
\usepackage[accsupp]{axessibility}  % Improves PDF readability for those with disabilities.

\usepackage{wrapfig}
\newcommand{\figref}[1]{Figure~\ref{#1}}
\newcommand{\tabref}[1]{Table~\ref{#1}}
\newcommand{\secref}[1]{Section~\ref{#1}}
\newcommand{\tabincell}[2]{\begin{tabular}{@{}#1@{}}#2\end{tabular}}  
\usepackage{stmaryrd}

\usepackage{marvosym}
\newcommand\blfootnote[1]{%	
  \begingroup
  \renewcommand\thefootnote{}\footnote{#1}%
  \addtocounter{footnote}{-1}%
  \endgroup
}

\begin{document}
% \renewcommand\thelinenumber{\color[rgb]{0.2,0.5,0.8}\normalfont\sffamily\scriptsize\arabic{linenumber}\color[rgb]{0,0,0}}
% \renewcommand\makeLineNumber {\hss\thelinenumber\ \hspace{6mm} \rlap{\hskip\textwidth\ \hspace{6.5mm}\thelinenumber}}
% \linenumbers
\pagestyle{headings}
\mainmatter
\def\ECCVSubNumber{3101}  % Insert your submission number here

\title{Temporal Saliency Query Network \\ for Efficient Video Recognition}

% INITIAL SUBMISSION 
\begin{comment}
\titlerunning{ECCV-22 submission ID \ECCVSubNumber} 
\authorrunning{ECCV-22 submission ID \ECCVSubNumber} 
\author{Anonymous ECCV submission}
\institute{Paper ID \ECCVSubNumber}
\end{comment}
%******************

% CAMERA READY SUBMISSION
% \begin{comment}
\titlerunning{Temporal Saliency Query Network for Efficient Video Recognition}
% If the paper title is too long for the running head, you can set
% an abbreviated paper title here
%

\author{
Boyang Xia$^{1,2}$\textsuperscript{*}\and
Zhihao Wang$^{1,2}$\textsuperscript{*}\and
Wenhao Wu$^{3,4}$\textsuperscript{\Letter}\and \\
Haoran Wang$^{4}$\and Jungong Han$^{5}$ \\
}

%  %将脚注符号设置为fnsymbol类型，即特殊符号表示
% \footnotetext[1]{These authors contributed equally to this work.} %对应脚注[1]
% \footnotetext[2]{Corresponding authors.} %对应脚注[2]

\authorrunning{Boyang Xia$^*$, Zhihao Wang$^*$, Wenhao Wu\textsuperscript{\Letter}, Haoran Wang, Jungong Han}
% First names are abbreviated in the running head.
% If there are more than two authors, 'et al.' is used.
%
% \institute{Key Lab of Intelligent Information Processing, Institute of Computing Technology, Chinese Academy of Sciences
%%- For short - %

\institute{Key Lab of Intelligent Information Processing of Chinese Academy of Sciences (CAS), Institute of Computing Technology, CAS, Beijing, China
\and  
University of Chinese Academy of Sciences, Beijing, China
\and 
The University of Sydney, Sydney, Australia
\and
Baidu Inc., Beijing, China
\and Computer Science Department, Aberystwyth University, SY23 3FL, UK
}
% \end{comment}
%******************
\maketitle

\begin{abstract}
Efficient video recognition is a hot-spot research topic with the explosive growth of multimedia data on the Internet and mobile devices. Most existing methods select the salient frames without awareness of the class-specific saliency scores, which neglect the implicit association between the saliency of frames and its belonging category. To alleviate this issue, we devise a novel Temporal Saliency Query (TSQ) mechanism, which introduces class-specific information to provide fine-grained cues for saliency measurement. Specifically, we model the class-specific saliency measuring process as a query-response task. For each category, the common pattern of it is employed as a query and the most salient frames are responded to it. Then, the calculated similarities are adopted as the frame saliency scores. To achieve it, we propose a \textbf{Temporal Saliency Query Network (TSQNet)} that includes two instantiations of the TSQ mechanism based on visual appearance similarities and textual event-object relations. Afterward, cross-modality interactions are imposed to promote the information exchange between them.
Finally, we use the class-specific saliencies of the most confident categories generated by two modalities to perform the selection of salient frames. Extensive experiments demonstrate the effectiveness of our method by achieving state-of-the-art results on ActivityNet, FCVID and Mini-Kinetics datasets. Our project page is at \url{https://lawrencexia2008.github.io/projects/tsqnet}.
\end{abstract}
\blfootnote{*: Co-first authorship. \Letter: Corresponding author.}

\section{Introduction}
In the recent years, video understanding has drawn considerable attention from the community~\cite{mamico,ASCNet,wu2021weakly,bcnet,Wu2022TransferringTK,wang2020symbiotic} for the inexorable increase of video content on the Internet. Much progress has been achieved on the techniques to model complex video events, which can be glimpsed on promising precision on multiple benchmark datasets \cite{kay2017kinetics,ucf101}. However, computational costs grow proportionally to the recognition accuracy. This hinders the deployment of video recognition systems in resource-constraint environments, \emph{e.g.} IoT, self-driving and mobile phone applications. Hence, it is imperative to develop efficient video recognition systems to meet the rising demands of resource-efficient applications.
% we provide a new perspective on action recognition by attaching importance to the semantic information of label texts rather than simply mapping them into numbers.Specifically, we model this task as a video-text matching problem within a multimodal learning framework, which strengthens the video representation with more semantic language supervision and enables our model to do zero-shot action recognition without any further labeled data or parameters requirements.

There are many studies that have been conducted on efficient video recognition.
One set of approaches focus on designing lightweight architectures \cite{r2plus1d,x3d}. 
% \emph{e.g.,} decomposing 3D convolution kernels to 2D and 1D ones \cite{r2plus1d} and progressively expanding 2D network to 3D one axis-by-axis \cite{x3d}. 
At the other end of the spectrum are the dynamic inference-based approaches, which typically utilize a lightweight policy network to preview the video events, and allocate computation resources depending on the saliency of frames. 
% which dynamically allocate resources on a per frame/clip basis, \emph{e.g.,} adaptively sampling salient frames \cite{scsampler19} and dynamically selecting frame resolutions \cite{arnet}. 
% The later follows the spirit of reducing redundancies in video content and easy to insert inside ever-advancing network architectures. 
They implant a policy network (or sampler network) inside the reinforce learning paradigm \cite{marl,adaframe,ada3d}, or adopt attention weight as a proxy of policy under the attention mechanism \cite{listentolook,smart2020}. The sampler networks are optimized under the assumption that the most salient frames/regions contribute most to  the video representation, which 
% observation1: 缺乏明确类别指向，容易误报
produces one-size-fits-all, \emph{i.e.,} class-agnostic frame saliency measurements. 
% class-agnostic

\begin{figure}[t] %H为当前位置，!htb为忽略美学标准，htbp为浮动图形
\centering %图片居中
\includegraphics[width=0.95\textwidth]{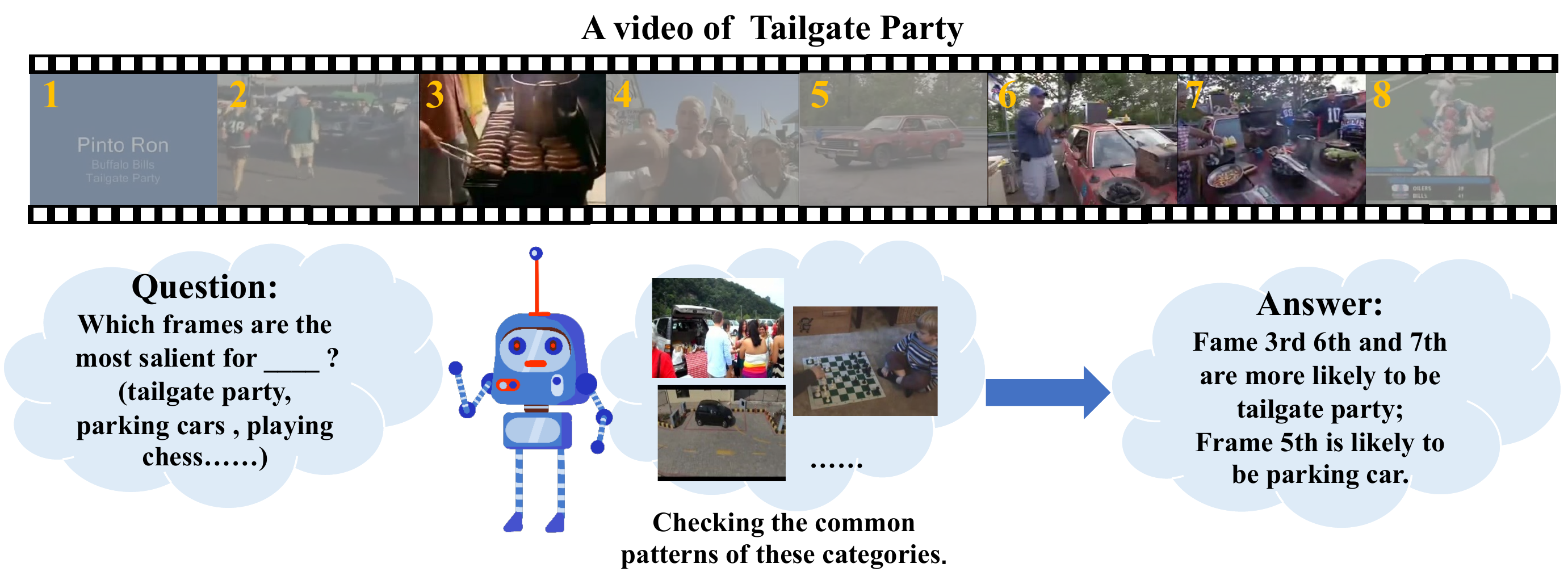} %插入图片，[]中设置图片大小，{}中是图片文件名
\caption{\textbf{A conceptual overview of the TSQ mechanism.} We cast the saliency estimation task as a query-response task. We ask each category a question: Which frames are the most salient ones for it? As we can see in the above example, we get the answer that frames 3rd, 6th, 7th are most salient for \emph{tailgate party} and the frame 5th is salient for \emph{parking car}. No frame is salient for \emph{playing chess}.}
\label{Fig.head} %用于文内引用的标签
\end{figure}

Actually, salient patterns are tightly associated with the category semantics. However, one-size-fits-all saliencies are not sensitive to fine-grained semantics. In particular, the sampler may overestimate the saliency of some frames which seem to be representative, but they actually belong to other categories rather than the real one of the current video. 
By contrast, a human can precisely elect the most informative frames with the aid of prior information about the probable category of the video. Because we can naturally build the logic connection between frame sequences and the common pattern of the predicted category, which can be understood as a query-response manner. For example, in \figref{Fig.head}, one can easily select the 3rd, 6th and 7th frames from the video with the assumption that the video may belongs to \textbf{Tailgate Party}. By contrast, one-sizes-fit-all sampler may also be inclined to 5th frames besides those three frames for it is quite representative for another category, \textit{e.g.} \textbf{Parking Car}.

%It is worth noting that since we do not know the category of this video in advance, we use all the categories to query this video. Therefore, when other categories appear in this video, there will also be a relatively large saliency score on the corresponding category response. For example, frame 5 in the figure is very similar to the action of Parking Cars. Therefore, the saliency score of this class will also be large, and we design Saliency Aggregation Module to solve this question.
% 这个现象lead us 来思考一个问题 如何在计算显著性的时候知道（引入）该视频的类别信息？我们通过假设该视频可能是任何类别来解决这个问题。
Inspired by this observation, in this paper, we cast frame saliency measuring as a querying process, to enable discriminative class-specific saliency measurement.
% to enable fine-grained discrimination power of the sampler.  
To this end, we present a novel \textbf{Temporal Saliency Query (TSQ) mechanism}, which can measure saliencies of all semantic categories over frame sequence in parallel, and select the saliency of highly-confident categories as final the result. 
% In TSQ mechanism, to comprehensively represent the common patterns of each category, we design meaningful TSQ embeddings from both visual frequent patterns and textual label semantics. 
Concretely, we formulate class-specific saliency measuring as a query-response task. The common patterns of the various categories are adopted as \textbf{query}, and frame representations gathered by category-frame similarities are taken as the \textbf{response}. Then, the category-frame similarities can be regarded as frame saliencies. A conceptual overview of the TSQ mechanism is shown in \figref{Fig.head}. Specifically, we use cross attention in Transformer Decoder \cite{transformer} to model many-to-many category-frame similarities in parallel. 
On one hand, we represent the common pattern of a category, namely TSQ embedding, by visual prototypes. And the query process is performed over the visual feature of the frame sequence. On the other hand, to handle large intra-class variations of visual appearance, we measure saliency by textual event-object relations for complementary information. As we know, the objects in videos are closely associated with the category annotation of video. For instance, \textbf{cake, candle and balloon with birthday party}. To model the semantic relationships between object and category, we first employ BERT \cite{bert} to represent the object with word embedding of its name. Taking the product as textual embedding, we construct another textual branch in the TSQ mechanism, where the query process is executed over the embedding sequence of object names. Doing so allows us to exploit prior knowledge from off-the-shelf word representations to supply cross-modal complementary clues to saliency measurement. 

\textbf{Our contributions are summarized as:}
\textit{First}, we propose a novel Temporal Saliency Query mechanism, to alleviate the lack of class-specific information in saliency measuring for temporal sampling frameworks. 
\textit{Second}, we present an efficient multi-modal salient frame sampler \textbf{T}emporal \textbf{S}aliency \textbf{Q}uery \textbf{Net}work\textbf{ (TSQNet)}, which utilize both visual appearance feature and textual feature obtained by object name embeddings to measure frame saliencies in a unified framework. \textit{Third}, we conduct extensive experiments on three large-scale datasets, \emph{i.e.,} ActivityNet, FCVID and Mini-Kinetics, which show TSQNet significantly outperforms the state-of-the-art approaches on accuracy-efficiency trade-off.

\section{Related Work}
\noindent\textbf{Efficient Video Recognition.} 
%%%% --- Related work of Video Recognition --- %%%%
% A substantial body of recent research have been devoted to recognize complex events or actions in videos by 2D CNNs \cite{twostream,tsn,tsm}, 3D CNNs \cite{c3d,i3d,s3d} or Transformers \cite{videoswintransformer}. 2D models apply 2D CNN on a per-frame basis and model temporal causalities by aggregation modules, \emph{e.g.,} pooling \cite{tsn} and channel shifting. 3D models process a input of stacked RGB frames with 3D convolutions and model spatial and temporal dependency jointly. Transformer based model use self-attention structures to model relations among spatial-temporal tokens \cite{transformer}. Although promising accuracy are achieved, limited studies focus on efficient video recognition \cite{x3d,mvf,adaframe} with favorable accuracy-cost trade-off. 
Efficient video recognition approaches can be roughly categorized into two directions. The first focus on elaborating new lightweight architectures by decomposing 3D convolution operations into 2D and 1D ones \cite{r2plus1d,s3d,mvf}, channel shifting in 2D CNNs \cite{tsm}, \emph{etc.} 
% and neural architecture search \cite{x3d,movinets}. 
% progressively expanding 2D network to 3D one axis-by-axis \cite{x3d}, \emph{etc.} 
The others are based on a dynamic inference mechanism~\cite{wu2020dynamic,nsnet,AKnet}, which allocates computation resources on a per-sample basis based on the saliencies of frames. 
% Several studies are conducted from different aspects of video content.  
% Korbar \emph{et al.} \cite{scsampler19} temporally sample salient clips from compressed video feature.
Wu \emph{et al.} \cite{marl} utilizes multi-agent reinforce learning to model parallel frame sampling and Lin \emph{et al.} \cite{ocsampler} make one-step decision with holistic view.
% Wu \emph{et al.} \cite{liteeval} dynamically decide to use either coarse CNN or fine CNN to process the frame 
Meng \emph{et al.} \cite{arnet} and Wang \emph{et al.}  \cite{adafocus,adafocusv2} focus their attention on spatial redundancy.
% select frame resolution adaptively to further reduce spatial redundancy. 
Panda \emph{et al.} adaptively decide modalities for video segments. Most of the previous works are mainly based on reinforce learning or attention mechanism, which are optimized with video classification objectives. However, this paradigm makes produced adaptive sampling policy class-agnostic and lacks discrimination power in fine-grained semantics. In contrast, our temporal sampling-based framework enables discriminative class-specific frame saliency measuring and shows that class-specific mechanism combined with visual-textual multi-modal complementary measuring can push the envelope of the trade-off between accuracy and computation cost.  

\noindent\textbf{Transformer in Vision Tasks.} Transformer \cite{transformer} is initially proposed to solve the long-term dependence problem in machine translation.
% It has been widely used in various tasks of natural language processing since it was proposed. 
ViT \cite{vit}, SwinTransformer  \cite{swintransformer} and DVT \cite{dvt} split image to patches as words and bring Transformer Encoder to computer vision classification tasks. Query2label \cite{query2label} apply Transformer Decoder to multi-label classification task.
DETR \cite{detr} explore using Transformer Decoder for object detection task.  
% which treats anchors as learnable embedding to query visual features for obtaining proposals. 
Then Transformer Decoder for segmentation is also developed by MaskFormer  \cite{maskformer}. 
The role of Transformer Encoder in C-Tran \cite{ctran} and TransVG \cite{transvg} is to model relations between different modalities.
% TQN \cite{tqn} adapt queries to ask fine-grained questions about event types and their attributes for fine-grained video classification.

%%%%% camera ready 觉得多余，删掉 %%%%%%%%%
% Although cross attention in Transformer Decoder plays an important role in our proposed method, there are still some differences between the above methods and ours.
% First, the purpose is different, 
% % they obtain the decoder outputs for downstream tasks and 
% we not only use decoder outputs but also take attention weights as salient scores.
% Second, the meanings of the query are different, their queries are randomly initialized, 
% % and all represent for some task-specific information 
% and our queries are the class-specific representation of categories which is initialized by common patterns/class name word embedding.
% % Third, compared with others methods which randomly initialize queries, we initialize queries using  to introduce class-specific category prior information and obtain better attention weights.
% % And the structure of corss attention fits perfectly with our class-specific querying motivation. 
%%%%% camera ready 觉得多余，删掉 %%%%%%%%%

\section{Method}
\begin{figure}[t] %H为当前位置，!htb为忽略美学标准，htbp为浮动图形
\centering %图片居中
\includegraphics[width=1.0\textwidth]{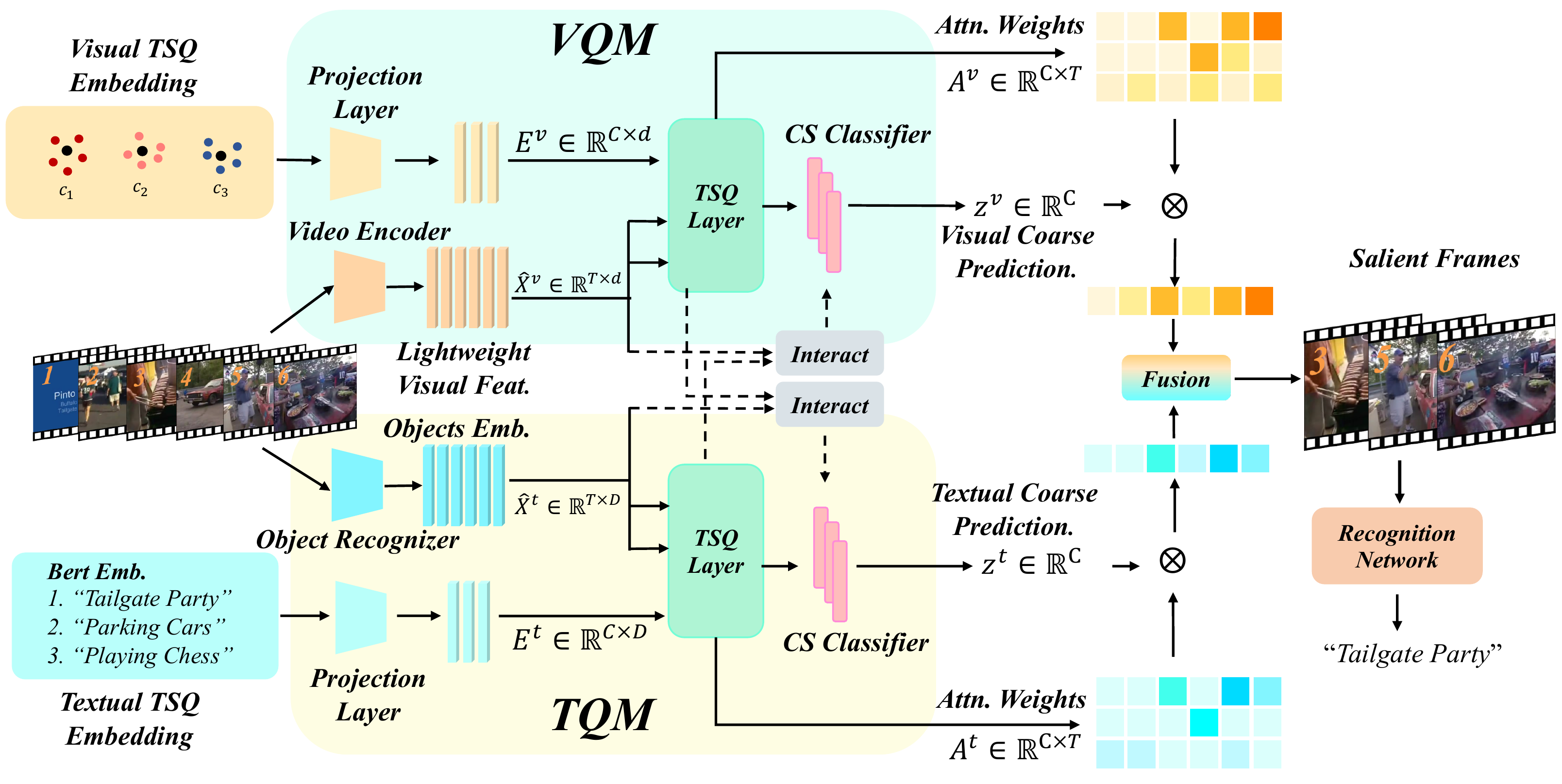} %插入图片，[]中设置图片大小，{}中是图片文件名
\caption{\textbf{The overview of the Temporal Saliency Query Networks.} Frame sequence is \textbf{queried} with visual and textual TSQ embeddings of categories in VQM and TQM, then TSQNet \textbf{responded} to the queries by gathering most salient frame representations for each category. And the resultant category-frame similarities are adopted as class-specific saliency measurements for two modalities, which are post-processed and fused for final saliency scores. Top $K$ frames with the highest saliency score are sampled and ingested to an off-the-shelf recognition network for final recognition. 
% the textual and visual extractor, and then do cross attention separately. Next, the classification results are obtained through the classification network. 
% Finally, we combined the classification result with attention weight to get the saliency score of each branch, then combined the of visual and textual branches to get the final frame saliency score, and used these saliency frames  for downstream video classification. 
% And for training, we add the Response Interaction Module for branch interaction which is shown using dotted line. 
Cross-modality interaction (``Interaction'') is considered for information exchanging during training. 
The projection layer is used to reduce the dimension of input features. 
} %最终文档中希望显示的图片标题
\label{Fig.main1} %用于文内引用的标签
\end{figure}

% 一些名字
% 分支名字 visual textual 参考VLbert visual query branch textual query branch
%  processing both visual and textual inputs in separate streams that interact through co-attentional transformer layers. 
Given a video of $T$ frames $X =\{x_i \}_{i=1}^{T}, x_i \in \mathbb{R}^{3\times H\times W}$ , our goal is to estimate the saliency score of frames $S =\{s_i \}_{i=1}^{T}$ and sample top $K$ frames with the highest saliency score to feed into a recognition network to obtain final video prediction $P$. The overview of our method is shown in \figref{Fig.main1}. 
% 这部分可能之后需要精简
% In this section, we first give an overview of our method in \secref{sec:overview}.
In this section, we first introduce the Temporal Saliency Query (TSQ) mechanism in \secref{tsq_mechanisim}. Then we elaborate on the framework of our TSQNet, including two instantiations of TSQ mechanism with visual and textual modalities and cross-modality interactions of them in \secref{TSQNet}. Finally, we present the inference procedure of TSQNet in \secref{inference}.
% how we aggregation the saliency scores together in \secref{sam}.

% Query response machimen,
% Multiclass response fusion module,
% response  Interaction module.

% \subsection{Overview}\label{sec:overview}
% Given a video of $T$ frames $X =\{x_i \}_{i=1}^{T}, x_i \in \mathbb{R}^{3\times H\times W}$ , our goal is to estimate the saliency score of frames $S =\{s_t \}_{i=1}^{T}$ and sample top $K$ frames with the highest saliency score to feed into a recognition network to obtain final video prediction $P$. 
% % 这部分可能之后需要精简
% The overview of our method is shown in \figref{Fig.main1}. 

% between textual TSQ embedding and the object-based frame-level embeddings 

\subsection{Temporal Saliency Query Mechanism}\label{tsq_mechanisim}
The goal of Temporal Saliency Query (TSQ) mechanism is to perform frame saliency estimation for all categories simultaneously, which is the shared building block for two branches of visual and textual modalities in TSQNet. To expand generic saliency to class-specific version, we are potentially to ask each category a question: 
% how similar is the content to the common pattern of each category? 
which frames are the most similar ones to the common pattern of it?
In this way, we can convert saliency estimation task to query-response task: a learnable embedding initialized with the common pattern of each category is set as the \textbf{query}, and the gathered feature from frame sequence with similarities is the \textbf{response}. Then the similarities between each category and frame sequence can be regarded as the saliency scores. We denote the learnable embedding here as \textbf{TSQ embedding}. In TSQ mechanism, a TSQ layer is proposed to enable the query-response functionality and a class-specific classifier is designed to generate coarce predictions of video category and enable discriminative learning of TSQ embedding for each category at the same time. The details of TSQ mechanism are described below.

\noindent\textbf{TSQ Layer.}
% \subsubsection{Attention}
The goal of TSQ layer is to model the many-to-many category-frame similarities simultaneously and enable learning of TSQ embeddings, denoted as $\{E_c \in \mathbb{R}^{d}\}_{c=1}^C$, under the video classification objective. To achieve this, TSQ layer is build on an attention structure in Transformer \cite{transformer}:
\begin{equation}
A_c = \operatorname{softmax}(\frac{Q_0 {K_0}^{\mathsf{T}}} {\sqrt{ d}}),~R_c=A_cV_0 ,
\end{equation}
% \begin{equation}
% R=A_cV
% \end{equation}\label{attention}
$Q_0\in \mathbb{R}^{d}$ is a query matrix, which is obtained by projecting each TSQ embedding $E_c$ with a parameter matrix $W_q \in \mathbb{R}^{d\times d}$: $Q_0 = E_cW_q$.
% , \emph{i.e.,} the common patterns of all $C$ categories. 
$K_0\in \mathbb{R}^{T \times d}$ and $V_0\in \mathbb{R}^{T \times d}$ are the key and value matrix, which are generated by projecting frame feature sequence $X\in \mathbb{R}^{T\times d}$ with different parameter matrices $W_k, W_v \in \mathbb{R}^{d\times d}$: $K_0=XW_k, V_0=XW_v$.
% $X $ are key and value matrix, which are embodied by frame feature sequence. $E_c$ is projected by a parameter matrix $W_q$, $X$ is projected by a parameter matrix $W_k$ and a parameter matrix $W_v$, respectively.
Then, for the TSQ embedding of the $c$-th category $E_c$, the attention weight $A_c \in \mathbb{R}^{T}$ is produced in querying process realized by scaled dot product operation. Then the value $V_0$ are gathered with attention weights $A_c$ and output as response vector $R_c\in \mathbb{R}^d$, which is fed to FFN of \cite{transformer}, \emph{i.e.,} sequential linear layers with residual connections. The output of FFN is ingested to a class-specific classifier to generate classification predictions. In addition to functioning as gathering weights, $A_c$ represent the frame saliency measurements of the $c$-th category for it characterizes the relations between the $c$-th category and all $T$ frames. In TSQ mechanism, the more discriminative $A_c$ is, the better the response vectors $\{R_c\}_{c=1}^C$ can represent the semantic information of the video, therefore the video classification objective can effectively optimize the this category-frame relation model.

\noindent\textbf{Class-specific Classifier.}
We denote the output of FFN as $\hat{R} \in \mathbb{R}^{C\times d}$ here. The goal of class-specific classifier (``CS Classifier'' in \figref{Fig.main1}) are twofold: (1) project $\hat{R} \in \mathbb{R}^{C\times d}$ to a coarse video prediction $z\in \mathbb{R}^{C}$, (2) enable class-specific learning of TSQ embeddings. In class-specific classifier, instead of directly using projection layer with weight matrix $W \in \mathbb{R}^{1\times d}$ as $z=W\hat{R}+b$, we apply $C$ projection layers with different weight matrices $\{W_c \in \mathbb{R}^{1\times d}\}_{c=1}^C$ to each $\hat{R_c}$ separately. For the $c$-th category, corresponding element of $p$ is computed as: 
\begin{equation}
z_c = W_{c} \hat{R}_{c}^{\mathsf{T}} + b_c,
\end{equation}\label{spfc}
where $b_c \in \mathbb{R}^{1}, b \in \mathbb{R}^{C}$ are the bias parameters (see Appendix for illustrative examples).
This class-specific design endows the response vector of each category with exclusive classifier, which effectively reserves the characteristic of each category and make model converge more easily. $z$ is used for calculating regular cross entropy loss with video labels. Notice here the difference between the coarse video prediction $z$ and the final video prediction $P$: $z$ is used for saliency measuring while $P$ is the final classification result of the recognition network.

% 讲一下 1d 卷积
% We first extract feature $F=\{f_i \}_{i=1}^{T}, f_i \in \mathbb{R}^{d}$ using $\Psi$ network.
% For a Transformer Decoder, it has two inputs, one as $Q$, the other as $K$ and $V$.
% Here we let learnable class label embedding to be $Q$, frame features $ F \in \mathbb{R}^{T, d}$ as $K$ and $V$. 
% We discarded the self-attention part in the Transformer Decoder and only keep the cross attention part.
% According to the Transformer Decoder structure above, class embedding $Q$ is send to self-attention to integrate information between categories.
% $$Q^{'} = Attention(Q,Q,Q)$$
% Then we use the feature $Q$ to query $F$ with corss-attention:
% $$Q^{'} = Attention(Q,F,F)$$
% Finally,$Q^{'}$ pass through $FFN$:
% \begin{equation}
% G_v = FFN_{v}(Q_{v}^{'}) \quad  
% G_t = FFN_{t}(Q_{t}^{'})
% \end{equation}\label{ffn}

\subsection{Temporal Saliency Query Network}\label{TSQNet}
Our TSQNet mainly consists of two modules: a Visual Query module and a Textual Query module, which are instantiations of TSQ mechanism with visual and textual representations, respectively. The Visual Query module query the frame appearance sequence with the visual TSQ embedding of each category, and collect the category-frame similarities for class-specific saliency estimation. Textual Query module measures saliencies by modeling event-object (or action-object) relations on the basis of prior knowledge in off-the-shelf language models. Besides, to exchange information between two TSQ modules, cross-modality interactions are performed synchronously during training, which effective compensate scarce scene information for Textual Query module. 
% The Saliency Generation Module produce final saliency measuring results by combining the salienceis of high-probability categories from two modalities selectively. We will elaborate on these modules in following sections.

\noindent\textbf{Visual Query Module (VQM).}
% \subsubsection{Visual Query Module (VQM).}\label{vqm}
The goal of VQM is to generate class-specific saliency measurement from pure visual perspective, which mainly consists of a video encoder, a TSQ layer and a class-specific classifier. The video encoder is a lightweight CNN or transformer backbone, \emph{e.g.,} MobileNetv2 \cite{mobilenetv2} and Mobile-former \cite{mobileformer}, which extract features from RGB frame sequence $\{x_i\}_{i=1}^T$ to feature sequence $\{\hat{x}_i^v \in \mathbb{R}^d \}_{i=1}^T$. We further use a 1D convolutional layer to reduce the feature dimension from $d$ to $d^{\prime}$, which we still denote as $\hat{X}^v=\{\hat{x}_i^v\}_{i=1}^T$ for brevity. TSQ layer takes visual TSQ embedding as query, and frame sequence as key and value, to generate saliency measurements $A^v\in\mathbb{R}^{C\times T}$ from visual features. Class-specific classifier produce visual video coarse predictions $z^v$, which is further used in the post-processing procedure of saliencies. Next we describe how we obtain visual TSQ embedding.

% \noindent\textbf{Visual TSQ Embedding.}
Following the definition in \secref{tsq_mechanisim}, visual TSQ embedding $\{E_c^v \in \mathbb{R}^d \}_{c=1}^C$ here is a set of learnable embeddings initialized with common appearance patterns of categories. We propose a simple prototype based representation for common appearance patterns here. Prior works \cite{proto_network} find that, most of the samples belonging to the same class cluster around a prototype in feature space formed by non-linear mapping of networks. We assume that category prototypes can represent the common patterns of categories. Following definitions in \cite{proto_network}, we use the averaged features of videos belonging to each category produced by video encoder in the training set, where a video feature is obtained by top-k pooling of frame features (see Appendix \ref{appendix:imp} for details).
A 1D convolutional layer is also used to project $E_c^v$ to the same $d^{\prime}$-dimension space with $\hat{x}_i^v$, which is still represented by $E_c^v$ hereafter.
% \begin{equation}
% Q_{v}^{'} = Attention_{v}(Q_v, \hat{X}_{v}, \hat{X}_{v})
% \end{equation}\label{visual_attention}

\noindent\textbf{Textual Query Module (TQM).}
% \subsubsection{Textual Query Module (TQM).}\label{tqm}
The goal of TQM is to provide knowledge-aware saliency estimation by mining generic event-object relations in videos with the help of prior knowledge in off-the-shelf language models. As observed by prior works \cite{15000object,smart2020}, the event-object (or action-object) relations are generic in videos. Although this knowledge is typically represented in knowledge graph \cite{image_text_matching}, we exploit it in a much more compact fashion, \emph{i.e.,} pre-trained language models.
% \emph{e.g.,} if cake, candle and balloon appears frequently in a video, the likelihood of the video belonging to \emph{birthday party} is quite high. 
It is proved that the semantic relationships between words can be effectively captured in pre-trained word representations, \emph{e.g.,} Word2Vec \cite{word2vec} and BERT \cite{bert}. To model category-frame relations, we first build a object vocabulary $W \in \mathbb{R}^{C_o\times D}$, on a pre-defined object list, \emph{e.g.,} ImageNet-1K category list ($C_o=1000$) with word embeddings. Then we introduce a lightweight but precise object recognizer to extract appearing object scores from each frame $\{O_i \in \mathbb{R}^{C_o} \}_{i=1}^T$. The frame-level object embedding based feature can obtained: $\hat{X}^{t}=\{\hat{x}_{i}^{t}\}_{i=1}^T, \hat{x}_{i}^{t}=O_i W$. Correspondingly, the textual TSQ embedding $\{E_c^t \in \mathbb{R}^D\}_{c=1}^{C}$ is initialized by pre-trained word embeddings of the category name, to align with textual feature sequence in embedding space. Similar to VQM, we add a 1D convolutional layer to $\{E^{t}_{c}\in \mathbb{R}^D\}_{c=1}^{C}$ and $\hat{X}^{t}$ to reduce dimensions, which are fed into a TSQ layer and class-specific classifier for textual frame saliency measurements $A^t \in \mathbb{R}^{C\times T}$ and textual coarse video prediction $z^t\in \mathbb{R}^C$.

\noindent\textbf{Cross-modality Interaction.}
% \subsubsection{Cross-modality Interaction.}\label{ci}
% \begin{wrapfigure}{r}{0.4\textwidth}%靠文字内容的右侧
% \centering
% % \includegraphics[width=0.4\textwidth]{intract.png}
% \includegraphics[width=0.3\textwidth]{intract.pdf}
% \caption{Illustration of Cross-modality Interaction Module, where residual connections are omitted. $A_v$ denotes the attention weights of visual branch, $A_t$ denotes the attention weights of textual branch. $V$ and $K$ denotes the textual feature sequence, $Q$ denotes \emph{category embedding} here.}\label{interact}
% \end{wrapfigure}
Here we seek to enable information exchange between TSQ layers of two modalities during training and provide guidance, \emph{e.g.,} scene knowledge, from VQM to TQM. To achieve this, we design a novel \textit{swap-attention} structure, which gather the feature sequence with attention weights of the other modality in both VQM and TQM, to generate two additional response vectors: 
% $\{R_c^{t\shortrightarrow v}\}_{c=1}^C$ and 
% $\{R_c^{v\shortrightarrow t}\}_{c=1}^C$, 
\begin{equation}
R^{t\shortrightarrow v}=A^t \hat{X}^v, R^{v\shortrightarrow t} = A^v \hat{X}^t ,   
\end{equation}
Then the two response vectors based on visual feature sequence $R^{v}$ and $R^{t\shortrightarrow v}$ are ingested to subsequent layers and compute loss as $\mathcal{L}_v$ and $\mathcal{L}_{t\shortrightarrow v}$. The same process conducted on textual features sequence renders $\mathcal{L}_t$ and $\mathcal{L}_{v\shortrightarrow t}$. 
% is illustrated in \figref{interact}. 
The swap-attention structure is conducive to TQM in two ways: (1) $\mathcal{L}_{t\shortrightarrow v}$ help optimize scene-aware category-frame relation model (2) $\mathcal{L}_{v\shortrightarrow t}$ help optimize scene-aware FFN and classifier. 
% : (1) textual attention weights $A^t$ learn 
% Scene information is beneficial for discrimination of many events \cite{wu2016harnessing}.
% The main motivation for the interaction module is using the attention weight of one branch assists the other branch. We added two new guided branches in our architecture. 
% From the previous introduction of cross attention, we can know that attention weight is a very important intermediate variable, which determines how to weight the frame-level information of a video. Therefore, a good attention weight will have a better guiding significance.
% Fig2 shows how we use visual attention weight guide textual branch. The original visual and textual cross attention are on the left and right, and the middle branch is composed of the visual attention weight combined with the V of the textual. On the other hand, we also design a branch of textual guided visual.
% \begin{equation}
% Q^{\prime}_{vt} = Attention(Q_{v}, K_{v}, V_{t})
% \end{equation}
% \begin{equation}
% Q^{\prime}_{tv} = Attention(Q_{t}, K_{t}, V_{v})
% \end{equation}\label{vt_tv}
% \begin{equation}
% Q^{\prime}_{vt} = \operatorname{CrossModalAttn}(Q_{v}, K_{v}, V_{t}) 
% \end{equation}
% \begin{equation}
% Q^{\prime}_{tv} = \operatorname{CrossModalAttn}(Q_{t}, K_{t}, V_{v})
% \end{equation}\label{vt_tv}
% The guidance branch brings in two new classification results and two loss functions.
We weighted the existing four losses to obtain the final loss function:
\begin{equation}
\mathcal L =  \mathcal L_{v} + \mathcal L_{t} + \alpha \mathcal L_{t\shortrightarrow v} + \beta \mathcal L_{v \shortrightarrow t} ,
\end{equation}\label{loss}
% Discussion of $\alpha$ and $\beta$ hyperparameters can be found in the supplementary materials.

\subsection{Inference of TSQNet.}\label{inference}
During inference, to yield final saliency measurements, we  aggregate the generated frame saliency estimation of high-probability predicted categories for two modalities, respectively, and fuse them for final saliency results. 
% Next we will decribe how to aggregate the attention weight of cross attention in each branch and how to merge two branches.

\noindent \textbf{Saliency Aggregation.}
Here we only describe saliency aggregation for VQM, which is conducted for TQM with the same way. Intuitively, the higher the probability that a video belongs in $c$-th category, the higher the priority of the $c$-th row of attention weights in final saliency result. Following this intuition, we aggregate class-specific saliency measurements of VQM, $A^v \in \mathbb{R}^{C\times T}$ with the coarse video prediction $z^v \in \mathbb{R}^{C}$. For the $i$-th frame, the measured saliency of VQM is:
\begin{equation}
s^v_i = \sum_{c=1}^{C}z^v_c A^v_{c,i},
\end{equation}\label{reweight1}
% re-weight $A$, 
% highest five categories,
% because different rows represent the attention scores of every frame with different categories. 
In practive, to filter the noise brought about by the low-confidence categories, we only aggregate saliencies of top-5 categories with highest $z^v$ to get final saliency measurements.
% And then we do mean operator for every frame among five categories. 
% Mathematically, final $s_t$ can be calculated as follows:
% \begin{equation} % y 是 200行5列，\mathbbm{1}是5行1列，p是200行1列
% y = \underset{y\in \{0,1\}^{C\times U}}{\operatorname{argmax}}\left< y, p \mathbbm{1}^\top \right>,
% % \quad \hat{p} = p \cdot y
% \quad \hat{p} = y \mathbbm{1},
% \end{equation}\label{argmax}
% where $p \mathbbm{1}^\top \in \mathbb{R}^{C\times U}$ are prediction logits replicated $U$ times and $\hat{p}$ is the one-hot representation of top-$U$ indices among $C$ categories.
% \begin{equation}
% s_t = \sum_{c=1}^{C}\hat{p}_c a_{c,t},
% \end{equation}\label{reweight2}
% % \begin{equation}
% % s_j = \frac{1}{5} \sum_{i=1}^{5}(A[index[i]][j])
% % s_t = \sum_{c=1}^{C}(A[c][t]\Tilde{y}[c])
% % \end{equation}\label{reweight}
% where the $U$ we use in our model is 5 here.

% Finally, we use the top-$K$ frames with the highest scores in $S= \{ s_t \}_{i=1}^{T}$ as the result of frame selection. The selected frames are feed into  computationally expensive model to do action recognition.
\noindent\textbf{Multi-modality Saliency Fusion.}
We fuse the saliency measurements of VQM and TQM by taking the union of the top $s_i^v$ frames and top $s_i^t$ frames. The number of frames used for union in two modules are controlled by pre-defined proportion $\lambda_v$ and $\lambda_t$, and the budget of selected frames $K$.

\section{Experiments}
% In this section, we conduct sufficient experiments on several large-scale action classification datasets to evaluate our method. In \secref{exp:setup} we present experimental details of our method. In \secref{exp:baseline} we compare the proposed method with some simples baselines. And the state-of-the-arts comparisons are presented in \secref{exp:sota}. In \secref{4.3}, we do extensive ablation studies to show the effectiveness of our modules. Finally, we make some qualitative analysis in \secref{exp:qualitative}.
\subsection{Experimental Setup}
\label{exp:setup}
\noindent \textbf{Datasets.}
We evaluate our method on three large-scale datasets: ActivityNet, FCVID and Mini-Kinetics.  
ActivityNet~\cite{caba2015activitynet} contains 200 categories, it has 10024 videos for training and 4926 videos for validation, where the average duration of videos is 117 seconds.
FCVID~\cite{fcvid} includes 91,223 videos which 45,611 for training and 45,612 for validation and divided to 239 classes, where the average duration of the videos is 167 seconds.
Mini-Kinetics is a small version of Kinetics \cite{kay2017kinetics}, it consists of 121k training videos and 10k validation ones from 200 categories. Different from first two benchmarks, the videos in Mini-Kinetics are trimmed, with a average length of 10 seconds.

\noindent \textbf{Evaluation metrics.}
For all datasets above, we apply the official train-val split to experiment our method.
Following the previous work, mean Average Precision (mAP) is used as the main evaluation metric for ActivityNet and FCVID, and Top1 accuracy for Mini-Kinetics. We also evaluate the computation cost with giga floating point operations (GFLOPs). 
%Computation cost is mainly affected by the network structure. 
% As a reference, the commonly used ResNet-50 and MobileNetv2 have 4.12 and 0.313 GFLOPS when input image is scales as 224 $\times$ 224.

\noindent \textbf{Implementation details.}
We adopt MobileNetv2~\cite{mobilenetv2} trained on target datasets as the video encoder in VQM, and Efficientnet-B0~\cite{efficientnet} trained on ImageNet-1K as the object recognizer in TQM, respectively. 
% $\Psi_{v}$ and $\Psi_{t}$. 
%%%% --- 后面再看是否需要这么细致 --- %%%% 
% For MobileNetv2, we used ImageNet pretrained models and then fine-tuned 10 epochs on the corresponding experimental dataset. 
% And for Efficientnet-B0 we use the model initialized on ImageNet directly.
% ResNet~\cite{resnet} series architectures are used as recognition networks.
% $\Phi$.
For fair comparisons with previous works, we adopt three backbones in ResNet \cite{resnet} series,  \emph{e.g.,} ResNet-50, 101, 152 for recognition networks. 
% All backbones are pre-trained on ImageNet and fine-tuned on corresponding dataset.
For resolution of frame processed by recognition networks, we follow previous works to scale the shorter side of frames to 256 and then center cropped them to $224\times224$ for all datasets. On ActivityNet and FCVID, the resolution of frames processed by VQM is $188\times188$ and one for TQM is $112\times112$ \footnotemark[1].
\footnotetext[1]{Note that the total computation cost of a $188\times188$ frame processed by MobileNetv2 and a $112\times112$ frame processed by EfficientNet-B0 equals to the cost of a $224\times224$ frame processed by MobileNetv2, which is the common setting of previous works \cite{adafocus,smart2020}.}
On Mini-Kinetics, the resolution is 112$\times$112 for both VQM and TQM.
\tabref{tab:flops} shows decomposition of computation cost of TSQNet when adopting ResNet-50 as recognition network. Please refer to Appendix for more implementation details.
% other dataset aug 
% For FCVID, we crop and resize to $224\times224$ following \cite{liteeval}.

%% 删除放在了 Appendix
\begin{comment}
We uniformly pre-sample $T$ frames from a video, and for those videos whose lengths are shorter than $T$, we repeat multiple times to padding it to $T$ frames.
% $T$ is set to 100 in ActivityNet and 129 in FCVID. 
Our frame sampler will select top $K$ most salient frames in $T$, 
% and feed them into expensive  network $\Phi$ to yield video-level prediction.
$T$ and $K$ can be adjusted to accommodate different budgets for downstream applications. 
We use SGD optimizer with momentum of 0.9 and train model with batch size of 64 for 100 epochs. The learning rate is  $10^{-2}$, decayed by the factor of 0.1 at the 25, 50, 75 epoch. Loss ratio $\alpha$ and $\beta$ are both 0.6. Fusion proportion $\lambda_v$ and $\lambda_t$ are 0.6 and 0.4, respectively.
\end{comment}

%%% --- SUPP. ---  %%%
% We use SGD optimizer with momentum of 0.9 and learning rate of  $10^{-2}$ for all experiments. Our learning rate is set to decay by the factor of 0.1 at the 25, 50, 75 epoch. The total training epoch of all our experiments is set to 100 and batch size is set to 64. 
\begin{table*}[t]
\centering
\begin{minipage}[t]{0.45\linewidth}
\caption{Example of FLOPs computation.}
\label{tab:flops}
\centering
\scalebox{0.8}{
\renewcommand{\arraystretch}{0.9}
\begin{tabular}{cccccc}
\toprule
Module & Arch.        & Res.             & FLOPs/F & \#F & FLOPs  \\
\midrule
% Previous Method & MobileNetv2     &    224     & 0.313G   & 16  & 5.01  G  \\
% \midrule
Vis.Enc. & MBv2     &    188     & 0.220G   & 16  & 3.52G  \\
Obj.Rec.  & EN-B0     &    112     & 0.098G   & 16  & 1.56G  \\
Rec.Net. & RN50       &    224    & 4.109G   & 5   & 20.55G \\
VQM & - & - &  -   & -   & 0.36G  \\
TQM & - & - &  -   & -   & 0.10G  \\
\midrule
Total   & -        &    -    & -       & -   & 26.09G 
\\ \bottomrule
\end{tabular}
}
\end{minipage}
\hfill
\begin{minipage}[t]{0.45\linewidth}
\caption{Comparisons with simple baselines.}
\label{table:baseline}
\centering
\setlength{\tabcolsep}{4pt}
\scalebox{0.8}{
\renewcommand{\arraystretch}{0.9}
\begin{tabular}{ccc}
\toprule
Method & mAP (\%) & FLOPs \\
\noalign{\smallskip}
\hline
\noalign{\smallskip}
Uniform &  70.9 & 195.8G \\
Random &  70.2 & 195.8G  \\
Dense &  71.2 &  930.8G \\
MaxConf &  74.2 & 930.8G \\
MaxConf-L & 71.2 & 54.9G \\
% Top K 10&  74.7\% & 930.8G \\
\midrule
% \textbf{Ours 50 10} &  \textbf{75.0} & \textbf{94.5G} \\
Ours &  74.3 & 55.3G \\
\bottomrule
\end{tabular}
}
\end{minipage}
\end{table*}
\subsection{Comparison with Simple Baselines}\label{exp:baseline}
We compare our TSQNet with some simple baselines with ResNet-101 without TSN-style training as the recognizer in \tabref{table:baseline}. There are multiple rule based baselines, ``uniform'' and ``random'' stand for uniformly and randomly selecting 10 frames from a video. ``Dense'' means using all frames of a video. For ``MaxConf'', we firstly obtain the maximum confidence among all categories for every frame by applying the model along time axis, then select $K$ frames with highest maximum confidence. We also compare with a simple sampler based baseline, ``MaxConf-L'', which is a lightweight version of ``MaxConf'' within a uniformly pre-sampled $T$ frames, as the same as ``ours''. The $T$ in ``MaxConf-L'' and ``ours'' is 50, and $K$ in ``MaxConf'', ``MaxConf-L'' and ``ours'' is 5.
 Our TSQNet obviously presents the best accuracy with limited FLOPs. In fact, ``MaxConf-L'' is an ablated baseline for our class-specific motivation, which replaces our TSQ mechanism with direct frame-level classification. Comparison with ``MaxConf-L'' confirms the efficacy of our TSQ mechanism.
%  each frame to get saliency score cross frames and categories. 
% The experimental results show that our method outperform these rule based methods and the imports of Temporal Saliency Query Mechanism.

\subsection{Comparison with State-of-the-arts}
\label{exp:sota}

\subsubsection{Results on AcitivtyNet.}
\begin{wrapfigure}{r}{0.5\textwidth} %H为当前位置，!htb为忽略美学标准，htbp为浮动图形
\centering %图片居中
\includegraphics[width=0.45\textwidth]{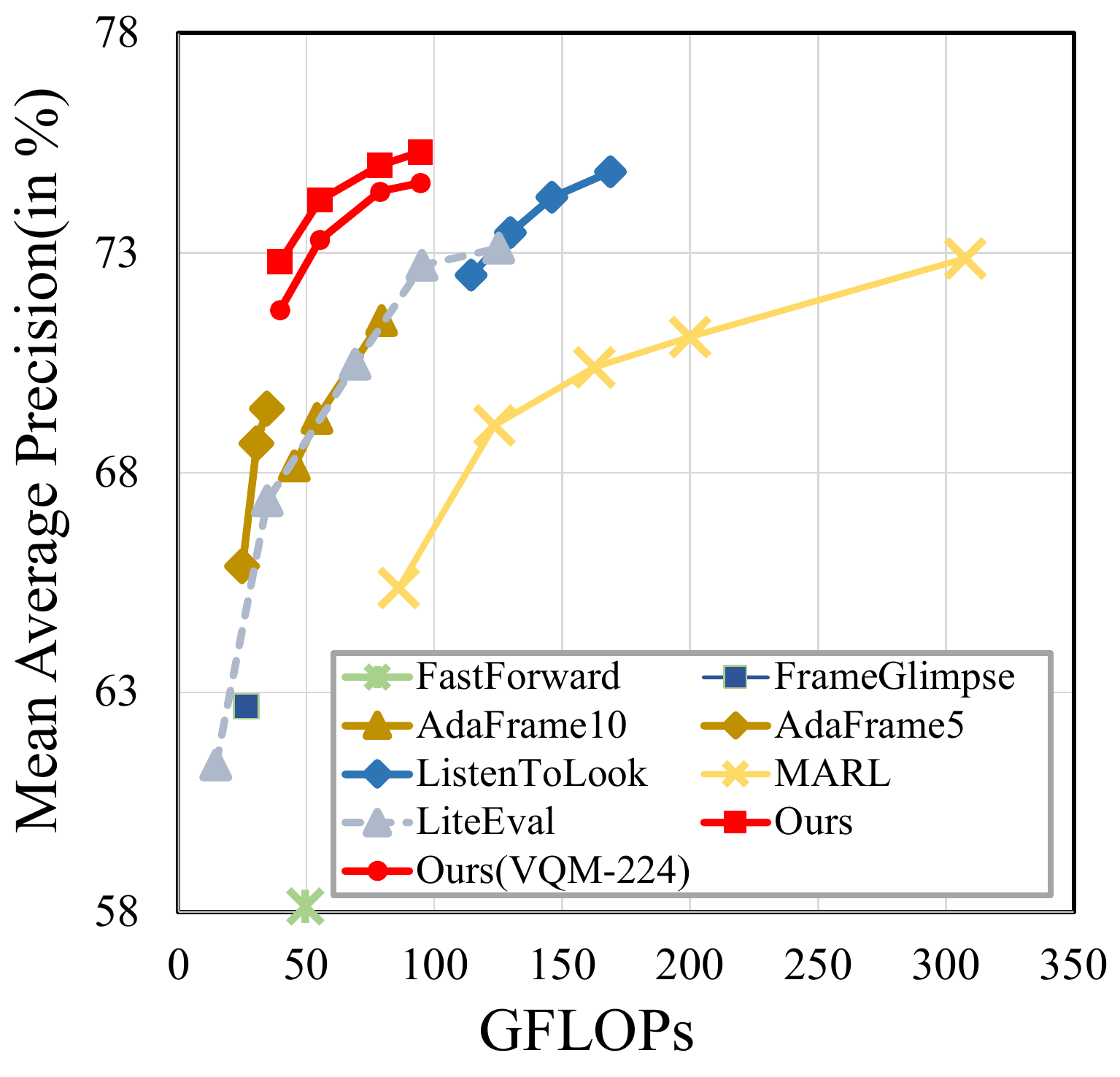} %插入图片，[]中设置图片大小，{}中是图片文件名
\caption{Comparison of the pure VQM and the whole TSQNet with the state-of-the-art based on ResNet-101 recognition network on ActivityNet.} %最终文档中希望显示的图片标题
\label{Fig.anet_res101} %用于文内引用的标签
\end{wrapfigure}
We compare the proposed method with recent SOTA methods on AcitivtyNet in \tabref{table:aet_sota_res50}: SCSampler~\cite{scsampler19}, AR-Net~\cite{arnet}, AdaMML~\cite{adamml}, VideoIQ~\cite{videoiq}, AdaFcous~\cite{adafocus}, Dynamic-STE~\cite{dynamicSTE} and FrameExit~\cite{frameexit}.
% And we report mAP, flops and the backbone of the recognition network used in all methods.
% For other methods, for other methods we directly use the reported results in the papers.
Experimental result shows that our method 
% achieves 76.5\% mAP on ActivityNet which
outperforms all existing methods with ResNet50 as the main recognition network. 
Compared with SCSampler~\cite{scsampler19} which is also a temporal sampling approach, our method surpass it by 3.7\% while using 1.6$\times$ less computation overhead, which demonstrates the discrimination power of TSQ mechanism in temporal saliency estimation.
% Compare with AdaFocus~\cite{adafocus} which is motivated by selecting salient spatial regions, we achieve 2\% higher (\textbf{76.5\%} v.s. 75.0\%) with less computation (\textbf{26.1G} v.s. 26.6G), which proves that our discriminative temporal sampler can capture more salient parts of videos.
Comparing to the state-of-the-art method based on early exiting, FrameExit~\cite{frameexit}, we still outperforms it by 0.5\%, which shows our class-specific sampler can find more discriminative frames than this sequential early exiting framework. For a more fair comparison with above pure visual based methods, we also present the results of the visual variant of TSQNet, \emph{i.e.,} `VQM-only' with comparable computes. Although without text modality, it still surpass the SotA methods, which verify the superiority of our TSM mechanism.

We further compare TSQNet with SOTA approaches in \figref{Fig.anet_res101} based on Res101 backbone. Following previous works \cite{adaframe,liteeval,listentolook,marl}, ResNet-101 without TSN-style training is used as the recognizer, as the same as in \secref{exp:baseline}. We calculate mAP under different budget $K$, which varies from 3 to 10.
It is shown that our method achieves clearly superior efficiency-accuracy trade-off over all methods. And the result of pure VQM illustrates the efficiency of TSQ Mechanism.
% Especially compared with ListenToLook~\cite{listentolook}, which is a audio-visual multi-modal method, we clearly outperform them with much less computational cost.

% Please add the following required packages to your document preamble:
% \usepackage{multirow}
\begin{table}[t]
\caption{Comparisons with SOTA efficient video recognition methods with ResNet50 as recognition backbone on AcitivtyNet. 188 and 224 here represent resolutions.}
\label{table:aet_sota_res50}
\centering
\setlength{\tabcolsep}{8pt}
\scalebox{0.85}{
\begin{tabular}{ccccc}
\toprule
Method  & Backbone & mAP(\%)  & FLOPs\\
\midrule
SCSampler~\cite{scsampler19}  & ResNet50 & 72.9 & 42.0G \\
% AR-Net  & \scalebox{0.85}{\begin{tabular}{c}MBv2, Res18,\\34,50\end{tabular}}        & 73.8 & 33.5G  \\
AR-Net~\cite{arnet}  &  ResNet18,34,50      & 73.8 & 33.5G  \\
AdaMML~\cite{adamml}  &  ResNet50 & 73.9 & 94.0G \\
VideoIQ~\cite{videoiq}  & ResNet50      & 74.8 & 28.1G  \\
AdaFocus~\cite{adafocus}  &  ResNet50      & 75.0 & 26.6G  \\
Dynamic-STE~\cite{dynamicSTE} & ResNet18,50 & 75.9 & 30.5G \\
FrameExit~\cite{frameexit} & ResNet50   & 76.1 & 26.1G  \\
\midrule
% \textbf{Ours} &  ResNet50  & \textbf{76.5}& \textbf{26.1G} \\
Ours (VQM-only$^{188}$) &  ResNet50  & 75.7 & 24.3G \\
Ours (VQM-only$^{224}$) &  ResNet50  & 76.5 & 26.1G \\
\textbf{Ours} &  ResNet50  & \textbf{76.6} & 26.1G \\
\bottomrule
\end{tabular}
}
\end{table}
\begin{table}[t]
\caption{Comparisons with SOTA video recognition methods using ResNet-152 and more advanced recognition networks on AcitivtyNet.}
\label{table:anet_sota_res152}
\centering
\setlength{\tabcolsep}{5pt}
\scalebox{0.85}{
\begin{tabular}{ccccc}
\toprule
Method  & Backbone & Pretrain & Accuracy(\%)  & mAP(\%)\\
\midrule
% C3D\cite{c3d} & -     & Sports1M   & 65.8 & 67.7 \\
P3D~\cite{p3d} & ResNet-152     & ImageNet   & 75.1 & 78.9 \\
RRA~\cite{rra} & ResNet-152     & ImageNet   & 78.8 & 83.4 \\
MARL~\cite{marl}  & ResNet-152     & ImageNet  & 79.8 & 83.8 \\
% ListenToLook\cite{listentolook}  & ResNet-152     & ImageNet &  80.3 & 84.2  \\
% Ours   & MBv2     & Res152     & ImgNet   & 83.8     &    \\
\textbf{Ours}   &  ResNet-152     & ImageNet   &  \textbf{80.0}   & \textbf{85.2}    \\ 
\midrule
% DSN & R(2+1)D-34 & K400 & 78.5 & 83.5\\
% MARL & BN-Inception & K400 & 80.2 & 83.5\\
% SMART~\cite{smart2020}     & ResNet-152     & Kinetics   & - & 84.4  \\
ListenToLook~\cite{listentolook}     & R(2+1)D-152     & Kinetics   & - & 89.9  \\
MARL~\cite{marl}     & SEResNeXt152     & Kinetics   & - & 90.1  \\
% \textbf{Ours}       & ResNet-152     & Kinetics   & \textbf{?}   & \textbf{?}\\
Ours       & Swin-B     & Kinetics   & 84.7   & 91.2\\
\textbf{Ours}       & Swin-L     & Kinetics   & \textbf{88.7}   & \textbf{93.7}\\
\bottomrule
\end{tabular}
}
\end{table}

To verify that our TSQNet can collaborate with more backbones, we present experiment results with ResNet-152 and Swin-transformer \cite{swintransformer} family as recognition networks in \tabref{table:anet_sota_res152}. It is shown that our method outperforms all method with the same ResNet-152 backbones, and achieves absolute SOTA precision (88.7 Top-1 accuracy and 93.7 mAP) with Swin-Transformer architecture.
% Finally, we achieve the state-of-the-art
% Finally, we achieve 90.2\% mAP by Swin-B\cite{} recognition backbone pretrained on Kinetics.

% Therefore, we can conclude that Modeling salient frames selection taks as the Query-Response task and introducing  the textual information of the category can achieve less computation and higher recognition accuracy.

\noindent \textbf{Results on FCVID.}
To verify that performance promotion can be achieved on more untrimmed datasets, we also evaluate our method on FCVID in \tabref{table:fcvid}, which shows that our method outperforms competing methods in terms of accuracy while saving much computation cost. 
Compared with SOTA approach AdaFocus~\cite{adafocus}, which is motivated by selecting salient spatial regions, we achieve higher mAP with less computation, which implies that our discriminative temporal sampler can capture more salient information of videos.
% We see that the our method outperforms AdaFocus\cite{adafocus}, which is a current method of dynamic focus frame area, with small computation (\textbf{26.2} vs 26.6 GFLOPs).

% Please add the following required packages to your document preamble:
% \usepackage{multirow}

\begin{table*}[t]

\centering
\begin{minipage}[t]{0.45\textwidth}
\centering
\caption{Comparison with SOTA efficient video recognition methods on FCVID. TSQNet achieves the best mAP with significant computation savings. `188' and `224' are resolutions.}
\label{table:fcvid}
\renewcommand{\arraystretch}{1.0}
\scalebox{0.9}{
\begin{tabular}{ccc}
\toprule
Methods & mAP(\%) & FLOPs \\
\noalign{\smallskip}
\midrule
\noalign{\smallskip}
LiteEval~\cite{liteeval}  &  80.0  & 94.3G \\
AdaFrame~\cite{adaframe}  &  80.2 & 75.1G  \\
SCSampler~\cite{scsampler19} & 81.0 & 42.0G  \\
AR-Net~\cite{arnet} & 81.3 & 35.1G  \\
AdaFuse~\cite{adafuse} & 81.6 & 45.0G  \\
SMART~\cite{smart2020} & 82.1 & -  \\
VideoIQ~\cite{videoiq} & 82.7 & 27.0G  \\
% Dynamic-STE~\cite{dynamicSTE} & - & -  \\
% FrameExit~\cite{frameexit} & - & - \\
AdaFocus~\cite{adafocus} & 83.4 & 26.6G  \\
\midrule
% \textbf{Ours} &  \textbf{83.5} & \textbf{26.2}G \\
Ours (VQM-only$^{188}$) &  82.9 & \textbf{24.4G} \\
Ours (VQM-only$^{224}$) &  83.3 & 26.2G \\
\textbf{Ours} &  \textbf{83.5} & 26.2G \\
\bottomrule
\end{tabular}
}
\end{minipage}
\hfill
\begin{minipage}[t]{0.45\linewidth}
\centering
\caption{Comparison with state-of-the-art methods on Mini-Kinetics. TSQNet achieves the best Top-1 accuracy with comparable computation cost with the most efficient methods.}
\label{table:Mini_Kinetics}
\renewcommand{\arraystretch}{1.0}
\scalebox{0.9}{
\begin{tabular}{ccc}
\toprule
Methods & Top-1(\%) & FLOPs \\
\noalign{\smallskip}
\midrule
\noalign{\smallskip}
LiteEval~\cite{liteeval}  & 61.0 & 99.0G \\
% AdaFrame~\cite{adaframe}  & - & - \\
SCSampler~\cite{scsampler19}  & 70.8 & 42.0G \\
AR-Net~\cite{arnet}    & 71.7 & 32.0G \\
AdaFuse~\cite{adafuse} & 72.3 & 23.0G \\
% SMART~\cite{smart2020}  & - & - \\
VideoIQ~\cite{videoiq}  & 72.3 & 20.4G \\
Dynamic-STE~\cite{dynamicSTE}  & 72.7 & 18.3G \\
FrameExit~\cite{frameexit}  & 72.8 & 19.7G \\
AdaFocus~\cite{adafocus}  & 72.9 & 38.6G \\
\midrule
% \textbf{Ours} &  \textbf{73.2} & \textbf{19.7G} \\
Ours (VQM-only) &  72.9 & \textbf{18.1G} \\
\textbf{Ours} &  \textbf{73.2} & 19.7G \\
\bottomrule
\end{tabular}
}
\end{minipage} 
\end{table*}

\begin{table*}[t]
\centering
\begin{minipage}[t]{0.45\linewidth}
\caption{Effectiveness of Class-specific designs.}
\label{table:spfc}
\setlength{\tabcolsep}{4.0pt}
\centering
\scalebox{0.8}{
\renewcommand{\arraystretch}{1.3}
\setlength{\tabcolsep}{3.0pt}
\begin{tabular}{ccc}
\toprule
Attention & Classifier & mAP(\%) \\
\noalign{\smallskip}
\midrule
\noalign{\smallskip}
CA             & CA         & 74.0    \\
CS             & CA         & 68.7    \\
CS             & CS         & \textbf{74.7}  \\
\bottomrule
\end{tabular}
}
\end{minipage}
\hfill
\begin{minipage}[t]{0.45\linewidth}
\centering
\caption{Effectiveness of multi-modality fusion and interactions.}
\label{table:select}
\setlength{\tabcolsep}{3.0pt}
\renewcommand{\arraystretch}{1.0}
\scalebox{0.8}{
\begin{tabular}{ccccc}
\toprule
$\mathcal{L}_{t\shortrightarrow v}$ & $\mathcal{L}_{v\shortrightarrow t}$ & TQM & VQM & Ours \\
\noalign{\smallskip}
\midrule
\noalign{\smallskip}
 - & - &  72.0 & 74.6 & 74.9\\
 $\checkmark$ & - & 72.5 & 74.8  &  75.1 \\
 - & $\checkmark$ &  72.7 & 74.6 & 75.1 \\
% \midrule
 $\checkmark$ & $\checkmark$ & \textbf{73.1} & \textbf{74.8} & \textbf{75.3} \\
\bottomrule
\end{tabular}
}

\end{minipage} 
\\ \vskip 2mm
\begin{minipage}[t]{0.45\linewidth}
\caption{Results of different textual feature.}
\label{table:textual}
\centering
\setlength{\tabcolsep}{6.0pt}
\renewcommand{\arraystretch}{1.15}
\scalebox{0.8}{
     \begin{tabular}{ccc}
\toprule
 Method & Usage & mAP(\%) \\
 \midrule
  W2V & Top10 & 71.2 \\
  Glove & Top10 &  72.0 \\
    \midrule
  Bert & All & 71.4 \\
  Bert  & Top10 & \textbf{72.1}  \\
 \bottomrule
 \end{tabular}
}
\end{minipage}
\hfill
\begin{minipage}[t]{0.45\linewidth}
\centering
\caption{Impacts of initialization of TSQ embedding.}
\label{table:init}
\setlength{\tabcolsep}{3.0pt}
\renewcommand{\arraystretch}{1.15}
\scalebox{0.8}{
 \begin{tabular}{ccc}
 \toprule
 Branch & Init & mAP(\%) \\
 \midrule
\multirow{2}{*}{Vis}  & Random  & 73.8\\
 & Prototype &  \textbf{74.7}\\
\midrule
\multirow{2}{*}{Text}  & Random &  71.6\\
& Bert Emb.   & \textbf{72.1}\\
 \bottomrule
 \end{tabular}
}
\end{minipage}

\end{table*}

\noindent\textbf{Results on Mini-Kinetics.}
We further test the capability of TSQNet on a short trimmed video dataset \emph{i.e.,} Mini-Kinetics, which is more difficult to sample salient frames.
% Note that the average duration of videos in Mini-Kinetics is much shorter than that of previous datasets (10 sec \emph{v.s.} 100+ sec), 
 \tabref{table:Mini_Kinetics} demonstrates that our method achieves superior Top-1 accuracy (\textbf{73.2} \emph{v.s.} 72.9) with 2.0$\times$ less FLOPs than the state-of-the-art method \cite{adafocus}.

% Overall, our method consistently outperforms state-of-the-art approaches on different datasets considering both accuracy and FLOPs, which demonstrates our TSQNet is a strong framework for various video recognition situations. 
\noindent\textbf{Practical latency.}
We further conduct experiments of practical efficiency, which shows that our TSQNet significantly surpasses two state-of-the-art methods in inference latency, \emph{i.e.,} FrameExit \cite{frameexit} (9.8 videos/sec \emph{v.s.} \textbf{TSQNet 121.1} videos/sec) and AdaFocus \cite{adafocus} (73.8 videos/sec \emph{v.s.} \textbf{TSQNet 121.1} videos/sec) \footnotemark[1]. See Appendix \ref{appendix:speed} for more details.
\footnotetext[1]{Results are obtained on a NVIDIA 3090 GPU with an Intel Xeon E5-2650 v3 @ 2.30GHz CPU.}
% \footnotetext[1]{We test the speed of two methods by running the official code released by the authors. We evaluate the inference speed of all methods on a NVIDIA 3090 GPU with Intel(R) Xeon(R) CPU E5-2650 v3 @ 2.30GHz CPU.} 
% FrameExit \cite{frameexit} reduce computation cost by early stopping in temporal sequential prediction.
% AdaFocus \cite{adafocus} suppose that the existing methods are spatially redundant, so it only selects salient areas to classify for each frames. 

\subsection{Ablation Study}
\label{exp:ablation}
% 需要做的消融实验有
% 1. 双分支的消融 OK
% 2. 聚类中心和随机初始化 
% 3. 融合方式
% 4. res101 50 选 10 8 5 3 

% 可能可以延续的有 
% 5. groupfc的 
% 6. top1 3 5 的 
% 7. head 和 layer 的 (和计算量有关的) 
% \input{table/practical_speed}
% \ref{exp:pracitcal_speed}
In this section, we inspect different aspects of our proposed TSQNet. All ablations are completed on AcitivtyNet with ResNet-101 as recognition network.

\noindent \textbf{Effectiveness of Class-specific Designs.}
We investigate the effectiveness of our class-specific designs in TSQ mechanism. 
% There are two class-specific modules in our method. 
% One of them is Temporal Saliency Query Mechanism, and the other is Class-specific-classifier. 
% We conduct some experiments in visual branch to verify the impact of these modules in \tabref{table:spfc}.
\tabref{table:spfc} presents the results of class-specific (``CS'') version and class-agnostic (``CA'') version of both the attention structure and the classifier in VQM. 
% CA stands for CLSS-Agnostic and CS stands for class-specific.
For attention structure, the class-agnostic version refers to setting the size of visual TSQ embedding set $\{E_{c}^v\}_{c=1}^C$ to 1. Then generated attention weight $A^v\in \mathbb{R}^{1\times T}$ is directly used as saliency measurement. For the classifier, the class-agnostic version is to replace existing $C$-projection-layer classifier with a single-projection-layer one as aforementioned in \secref{tsq_mechanisim}. 
% because $Q \in \mathbb{R}^{1, d_v}$. 
% This implementation  corresponds to the first row in the \tabref{table:spfc}.
% The Class-specific-classifier module is also an important part which fits perfectly with our motivation.
% We also maintain Temporal Saliency Query Mechanism and just  replace the Class-specific-classifier module with a full connect layer which corresponds to the second row in the \tabref{table:spfc}.
It is shown that ``CS CS'' (ours) significantly outperforms ``CA CA'' choice, which confirms the effectiveness of class-specific information in saliency measurements. Besides, ``CS CA'' choice presents an unpromising result, which demonstrates that class-specific classifier is critical for TSQ mechanism to function normally in class-specific setting. See Appendix for illustrative examples of these three settings and detailed explanation of comparison of their performance.
% Experimental results demonstrate that class-specific designs in both attention structure and classifier play important roles in TSQNet. 
% we also did some analysis of the second row result. Compare to the traditional classification task which the features used for classification mix the information of all categories, 
% our features is class-specific. 
% Taken the AcitivtyNet  which has 200 categories as an example the extracted features are 200 times more than the general classification task and the features of each category are independent. Therefore, 200 classifiers is applied in our method. And only one full connect layer will not be able to classify all category.

\noindent\textbf{Effectiveness of Multi-modal and Fusion and Interactions.}
To verify the effectiveness of fusion of VQM and TQM and multi-modality interactions, we present experimental results on two individual modalities with different usage of $\mathcal{L}_{t \shortrightarrow v}$ and $\mathcal{L}_{v \shortrightarrow t}$ in \tabref{table:select}. 
% First of all, we can see that when the two branches conduct experiments independently.
% the result of the visual branch is XXX, the result of the textual branch is XXX, and the result of the fusion of the two branches is XXX.
Without any interactions, fusion of two modules relatively impart improvements on TQM and VQM for 2.9\% and 0.3\% respectively, which verifies that two modules are complementary. $\mathcal{L}_{t \shortrightarrow v}$ clearly elevate the performance of TQM for better category-frame modelling guided by visual features from VQM. The performance of VQM is also slightly improved by introducing textual-modality attention weights. $\mathcal{L}_{v \shortrightarrow t}$ significantly improves the performance of TQM for better learning of textual FFN and classifier.
% When we add the visual branch to guide the text branch, we can see that the result of the text branch is greatly improved, and the final fusion result is also improved.
Finally, when both losses in CIM are added, the results of both TQM and VQM branch are further promoted, and performance of overall TSQNet is obviously improved (\textbf{75.3} \emph{v.s.} 74.9). See Appendix \ref{appendix:ablation} for detailed investigations on ratios of two losses. 
% The ablation experiment results prove that the proposed text branch can complement the visual branch, and the interaction between the two branches can better promote the results of the two branches.

\noindent \textbf{Different Textual Feature.}
In \tabref{table:textual}, we try three commonly used word embeddings, \emph{i.e.,} Bert~\cite{bert}, Glove~\cite{glove} and Word2Vec~\cite{word2vec}, as well as two fashions of usage of object scores $O_i$, \emph{i.e.,} top-10 object categories (``Top10'') and all categories (``All'').
Experimental result shows that the Bert embedding with top-10 object score gain the best result, which verifies that both the quality of word embedding and noise filtering of object category count for textual instantiation of TSQ mechanism.
%%%%%%%%%%%%%%%%%%%%%%%%%%%%%%%%%%%%%%%%%%%%%%%%%%%%%%%%%%%%%%%%%%%
% 下面的是旧的注注释
% and  represent for two ways to obtain frame level textual features. One represents weighting the Bert embedding of 
% top10 ImageNet categories , and the other represents weighting all 1,000 categories.
% On the one hand, Bert embedding learns better semantic than other emneddings and Bert can better model the sequence semantics, On the other hand, using all categories will produce more noise, thus affecting the text information of the frame.
% In the first method, we only use the 10 categories with the highest classification scores, then use their scores to re-weight the Bert features corresponding to the names of these categories, and finally sum the 10 features as the features of the sample.For the second method, we directly weighted the class names of all 1000 classes as the final result. 
%%%%%%%%%%%%%%%%%%%%%%%%%%%%%%%%%%%%%%%%%%%%%%%%%%%%%%%%%%%%%%%5

\noindent\textbf{Impacts of Initialization of TSQ Embedding.}
We further explore the initialization of visual and textual TSQ embeddings in \tabref{table:init}. 
% In the visual branch, we used the category center for initialization; in the text branch, we used the bert feature of the category name for initialization.
The comparison with random initialization confirms that proposed prototype based visual TSQ embedding in VQM and word embedding based textual embedding in TQM provide meaningful and effective initialization for TSQ embeddings. 
% For more intuitive understanding of TSQ embeddings, we present t-SNE visualization of them in \secref{visualization}.
% at the start and end of training, please refer to  and supplementary material.
% In this section, we visualize the visual and textual TSQ embeddings by t-SNE in \figref{q_init} and show qualitative examples of sampled frames by TSQNet in \figref{qualitative_example}.
% This demonstrates that training with TSQ mechanism compensate textual TSQ embeddings for rich visual semantic information. 

%%% --- embedding可视化 --- %%%
\begin{figure}[!t] %H为当前位置，!htb为忽略美学标准，htbp为浮动图形
\centering %图片居中
\includegraphics[width=1.0\textwidth]{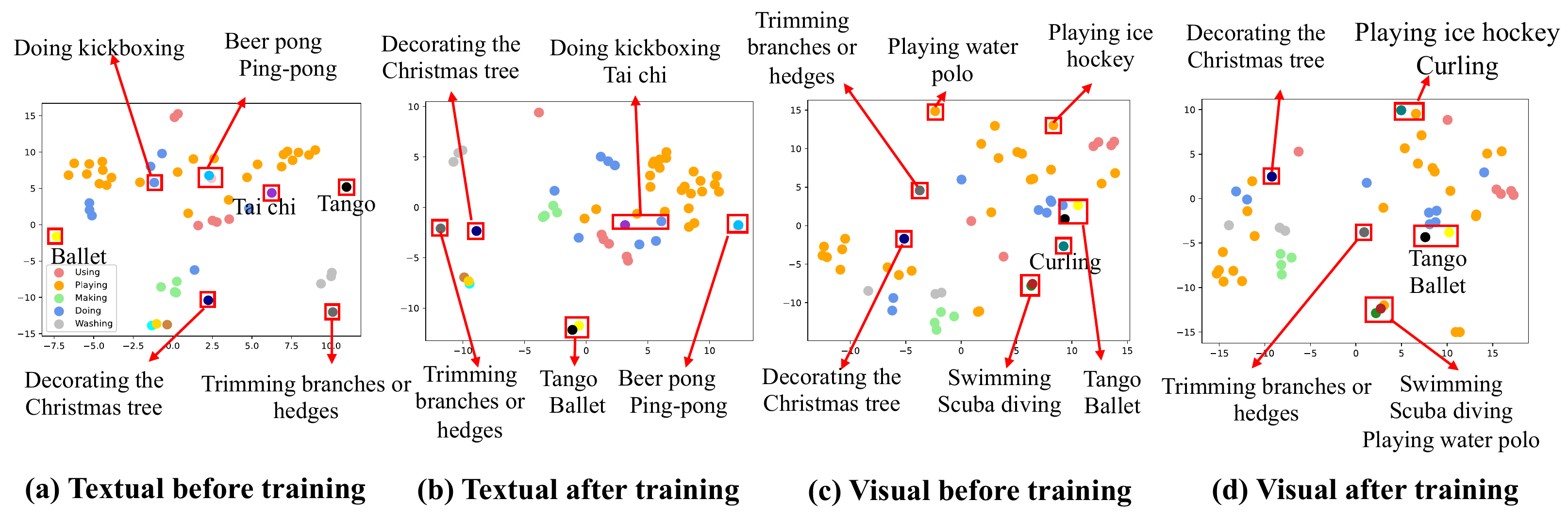} %插入图片，[]中设置图片大小，{}中是图片文件名
\caption{\textbf{The visual and textual TSQ embeddings before and after training visualized by t-SNE. }The category embeddings with relevant semantics cluster together after training. See \secref{exp:qualitative} for detailed explanation. 
% - 为了篇幅注释 - % 
% As shown in (a) and (b), textual TSQ embeddings of the categories sharing the same words, \emph{e.g.,} \textbf{Ping-pong} and \textbf{Beer pong}, cluster together before training because of the close relationships in Bert embedding space, which remain together till training is over. Some closely related categories, \emph{e.g.,} \textbf{Tango} and \textbf{Ballet}, with similar visual cues cannot cluster together in textual before training, while they cluster together after training. Similar phenomenon can be observed in visual TSQ embeddings in (c) and (d). \textbf{Playing Water Polo} joins the cluster of \textbf{Swimming} and \textbf{Scuba diving} after training, which implies visual semantics are enhanced during training of TSQ mechanism.
} %最终文档中希望显示的图片标题
\label{q_init} %用于文内引用的标签
\end{figure}
%%% --- embedding可视化 --- %%%
\begin{figure}[!t] %H为当前位置，!htb为忽略美学标准，htbp为浮动图形
\centering %图片居中
\includegraphics[width=0.9\textwidth]{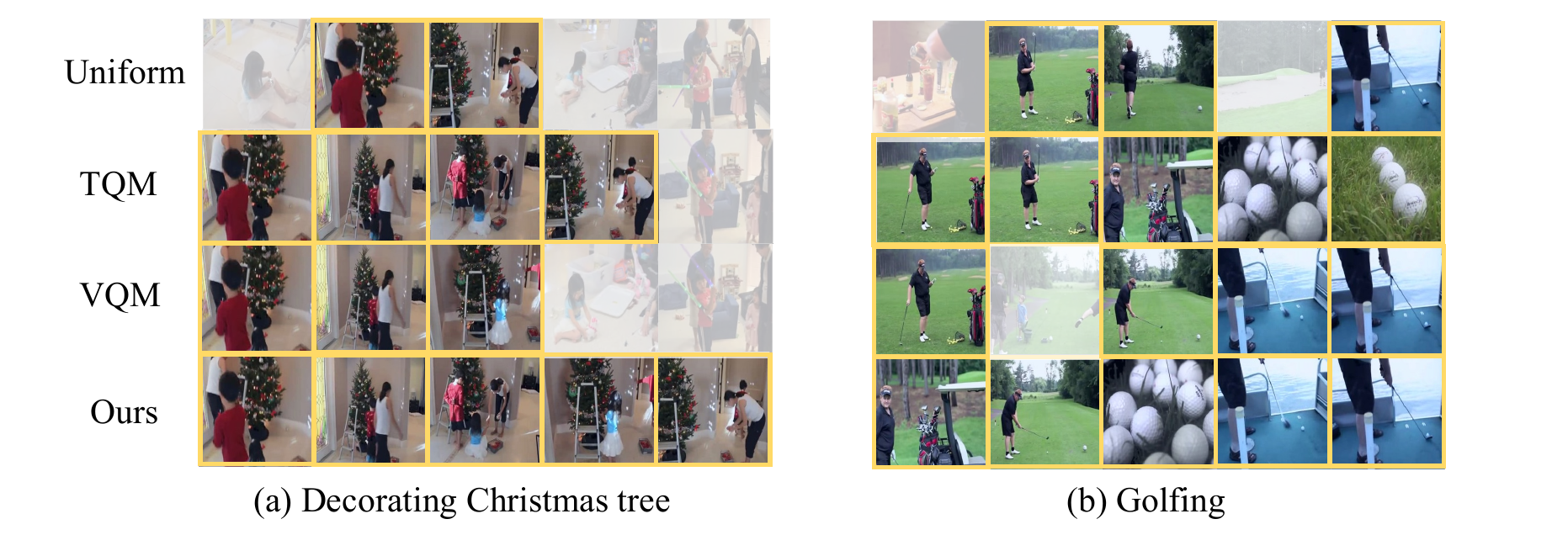} %插入图片，[]中设置图片大小，{}中是图片文件名
\caption{\textbf{Qualitative Evaluation of Sampled frames.} We visualized the most salient five frames of uniform and our proposed methods with two samples. The frames with \textcolor{Gold}{golden} border represent the identified salient frames by human intuition, and the frames with mask denote the non-salient ones. } %最终文档中希望显示的图片标题
\label{qualitative_example} %用于文内引用的标签
\end{figure}

% -- Practical Inference Speed -- %
% \subsection{Practical Inference Speed}\label{exp:speed}
% To further verify the practical efficiency of our method, we compare the inference speed with two state-of-the-art methods, \emph{viz.,} FrameExit \cite{frameexit} and AdaFocus \cite{adafocus} on ActivityNet \footnotemark[1]. \footnotetext[1]{We test the speed of two methods by running the official code released by the authors. We evaluate the inference speed of all methods on a NVIDIA 3090 GPU with Intel(R) Xeon(R) CPU E5-2650 v3 @ 2.30GHz CPU.} 
% % FrameExit \cite{frameexit} reduce computation cost by early stopping in temporal sequential prediction.
% % AdaFocus \cite{adafocus} suppose that the existing methods are spatially redundant, so it only selects salient areas to classify for each frames. 
% Results in two different settings with batch size of 1 and 32 are reported. 
% Note that FrameExit \cite{frameexit} exits from recognition at different time for different videos, so it cannot inference in batch setting, which we only report the latency with batch size = 1 here. Experimental results in \tabref{table:inference_speed} show that our TSQNet surpass other methods significantly in practical inference speed (\textbf{121.1} video/s).
% .
% 84.33152332276703
% 67.99111673069753

\subsection{Qualitative Analysis}\label{exp:qualitative}
% One more interesting thing is that the textual branch can obtain object information which visual can not. 
%%% --- embedding可视化 --- %%%
We visualize visual and textual TSQ embdding by t-SNE in \figref{q_init}, which shows that our class-specific motivation is highly interpretable in terms of relationships between categories.
We also find some categories sharing similar objects are more closer in text TSQ embeddings than in visual ones. For examples, \textbf{Decorating the Christmas tree} and \textbf{Trimming branches or hedges} share tree or tree-related objects and become closer after training. This may be because TQM measure saliency based on event-object relations, which are more robust against scene variations. 
%%% --- embedding可视化 --- %%%
% Because of the differences in decorations and scenes between Christmas trees and common trees, visual did not cluster them together at the end of the training.
In \figref{qualitative_example}, we exhibit some qualitative examples of \textbf{Decorating Christmas tree} and \textbf{Golfing} for sampled frames by uniform baseline, TQM, VQM and TSQNet. 
In the case of \textbf{Decorating Christmas tree}, it is shown that TQM and VQM are clearly better than uniform baseline. After fusion, TSQNet can sample further more salient frames.
Another qualitative example \textbf{Golfing} is quite interesting.
% We can see that the first image in uniform is irrelevant.
VQM captures the action moments of swinging a golf club and scenes of a golf course, while TQM captures the golf balls and a golf cart. 
After fusion, TSQNet select the frames of these object, actions and scenes,
% contain the golf ball and the swing of the golf club 
which implies our TQM and VQM can cooperate to build a robust sampler aware of object, scene and action information.

\section{Conclusions}
This paper investigates efficient video recognition by proposing a novel Temporal Saliency Query mechanism and presents an efficient multi-modal salient frame sampler Temporal Saliency Query Network.
Extensive experiments verify the proposed method significantly outperforms the state-of-the-art approaches on accuracy-efficiency trade-off.
% 这里还差一句话，展望外来或者表达我们的方法的可拓展性
Our proposed method is model-agnostic and can be used with various network architectures. And since our salient score is class-specific, we can easily extend our method to multi-label efficient video recognition.

\appendix
\section*{Appendix} \label{appendix}
\setcounter{table}{0}
\renewcommand{\thetable}{A.\arabic{table}}
\renewcommand{\thefigure}{A.\arabic{figure}}

\section{Further Implementation Details}\label{appendix:imp}
Here we provide some implementation details of TSQNet. 
We uniformly pre-sample $T$ frames from a video, and for those videos whose lengths are shorter than $T$, we repeat multiple times to padding it to $T$ frames.
% $T$ is set to 100 in ActivityNet and 129 in FCVID. 
Our frame sampler will select top $K$ most salient frames in $T$, 
% and feed them into expensive  network $\Phi$ to yield video-level prediction.
$T$ and $K$ can be adjusted to accommodate different budgets for downstream applications. 
% See Supp. for details of training. 
We use SGD optimizer with momentum of 0.9 and train model with batch size of 64 for 100 epochs. The learning rate is  $10^{-2}$, decayed by the factor of 0.1 at the 25, 50, 75 epoch. Loss ratio $\alpha$ and $\beta$ are both 0.6. Fusion proportion $\lambda_v$ and $\lambda_t$ are 0.6 and 0.4, respectively.
% there are altogether three backbone networks, which are Visual Encoder, Object recognizer and Recognition network which use backbones as follows:
We use MobileNetv2 and EfficientNet-B0 as the video encoder in VQM and object recognizer in TQM, respectively. 
For video encoder in TQM, we use the ImageNet pre-trained model and finetuned it on target datasets \textit{e.g.,} ActivityNet, \emph{etc.,} for 10 epochs. And for Object recognizer, we directly use the officially released ImageNet model to extract object score of the ImageNet 1000 classes. We use positional embedding  on frame sequence in transformer decoder to model temporal order information.
% Finally, for recognition networks, we do the same operation as video encoder but finetune for more epochs.

\noindent \textbf{Prototype feature generation.}
Here we introduce how we obtain visual prototype based representation for visual TSQ embeddings initialization. 
% First, we found the video samples corresponding to each category in the training set. 
First we apply a classifier to get the classification results for each frame. 
Then we select the top $m$ percent of frames which can correctly predict the ground truth video category, which are then averaged to obtain the representation of each video. 
Finally, we pool all the video representations of each category to get the prototype representation of each category.
We use $m = 30$ for all experiments in this paper.

\noindent \textbf{Saliency score fusion.}
We describe in detail how to fuse the VQM and TQM saliency scores into the final saliency measurement. Suppose we have the VQM salient scores $S^v \in \mathbb{R}^{T} $ and TQM salient scores $S^t \in \mathbb{R}^{T} $ of one video. We join the top saliency score frames from two modalities to get final $K$ salient frames. Specifically, the number of selected frames from two modalities are determined by  $\lambda_v K$ and $\lambda_t K$, respectively, where $\lambda_v + \lambda_t=1$.  
% We use $\lambda_v = 0.6 $ in all experiments. 
For example of selecting 5 frames from 16 frames with $\lambda_v=0.6$ situation, we select top $5 \times 0.6=3$ frames from VQM and top $5-3=2$ ones are from TQM. And if there exists duplication, which results in a final result of less than 5 frames, the selection will be deferred in the VQM according to the descending order of $S^v$ until meeting the 5-frame budget. We use $\lambda_v=0.6$ and $\lambda_t=0.4$ in experiments.

\section{Practical Inference Speed}\label{appendix:speed}
To further verify the practical efficiency of our method, we compare the inference speed with two state-of-the-art methods FrameExit \cite{frameexit} and AdaFocus \cite{adafocus} on ActivityNet. 
FrameExit \cite{frameexit} reduce computation cost by early stopping in temporal sequential prediction.
AdaFocus \cite{adafocus} suppose that the existing methods are spatially redundant, so it only selects salient areas to classify for each frames. We test the speed of two methods by running the official code released by the authors.
We evaluate the inference speed of all methods on a NVIDIA 3090 GPU with Intel(R) Xeon(R) CPU E5-2650 v3 @ 2.30GHz CPU. Results in two different settings with batch size of 1 and 32 are reported. 
Note that FrameExit \cite{frameexit} exits from recognition at different time for different videos, so it cannot inference in batch setting, which we only report the latency with batch size = 1 here. Experimental results in \tabref{table:inference_speed} show that our method not only saves much theoretical computation complexity but also achieves the fastest actual inference speed (\textbf{121.1} video/s) on both single-sample and batch setting.
% .
% 84.33152332276703
% 67.99111673069753
\begin{table}
\begin{center}
\caption{Comparisons of practical inference speed with state-of-the-art methods on ActivityNet.}
\label{table:inference_speed}
\begin{tabular}{ccccc}
\toprule
Method  & mAP (\% ) & FLOPs (G) & \tabincell{c}{Throughput(bs=1) \\ (videos/s)}   & \tabincell{c}{Throughput(bs=32)\\(videos/s)} \\
\midrule
AdaFocus \cite{adafocus} & 75.0 & 26.6 & 5.5 & 73.8 \\
FrameExit \cite{frameexit} &  76.1 & \textbf{26.1} & 9.8 & - \\
\textbf{Ours} & \textbf{76.5} & \textbf{26.1} & \textbf{17.7} & \textbf{121.1} \\
\bottomrule
\end{tabular}
\end{center}
\end{table}

\section{Additional Ablation Study}\label{appendix:ablation}
In this section, more ablation experiments are conducted to supplement the main paper. ResNet-101 is utilized for the recognition network as the same as in ablation studies of the main paper.

\subsection{Ablation of class-specific classifier}
\begin{figure}[h] %H为当前位置，!htb为忽略美学标准，htbp为浮动图形
\centering %图片居中
\includegraphics[width=0.7\textwidth]{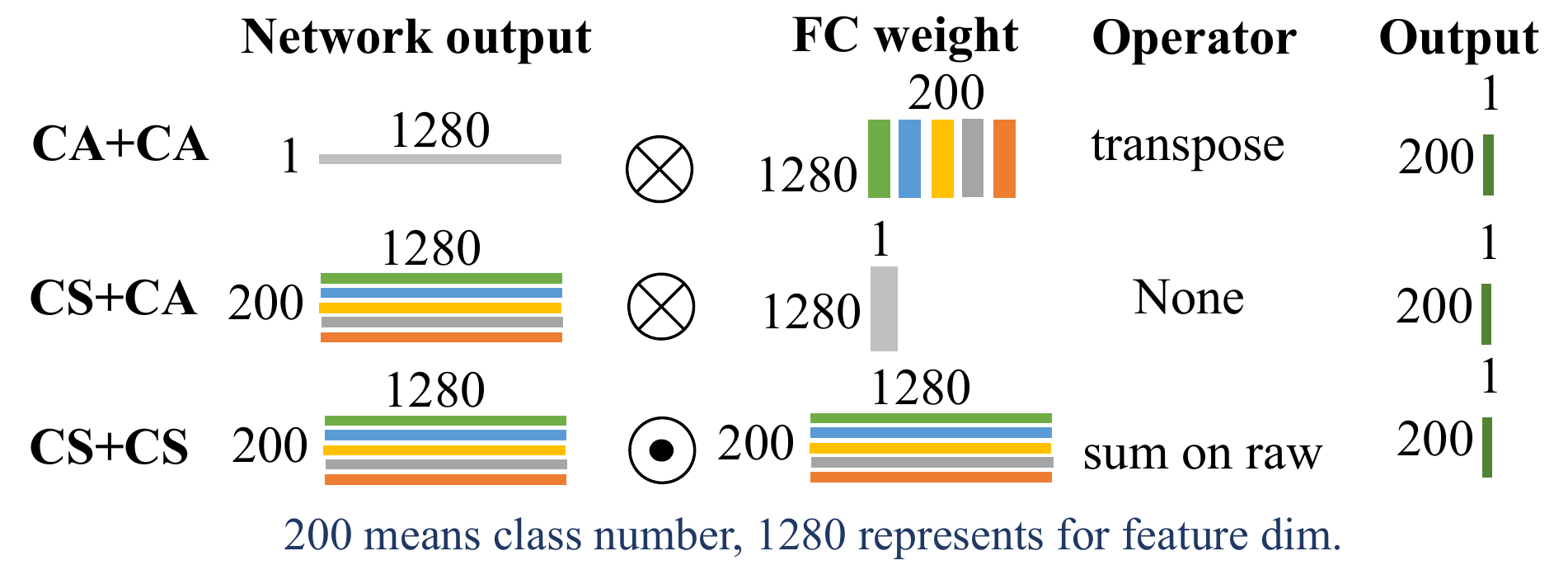} %插入图片，[]中设置图片大小，{}中是图片文件名
\caption{Illustrative examples of three combinations between the attention structure and the classifier of TSQNet, {i.e.} ``CS+CA'', ``CA+CA'' and ``CS+CS''.} %最终文档中希望显示的图片标题
%\label{Fig.head} %用于文内引用的标签

\end{figure}
% \vspace{-0.05cm}
% \\``CS+CA'' is an ablation of ``CS+CS''. 
We show the illustrative examples of the combinations of the attention structure and the class-specific classifier under the situation of 200 class and 1280 feature dimensions in Figure~\ref{Fig.head}. ``CA + CA'' represents the class-agnostic attention structure (with 1 query) combined with the class-agnostic classifier (with a single FC). ``CS + CA'', {i.e.} our TSQNet, the class-specific attention structure (with $C$ queries) combined with the class-agnostic classifier (with a single FC). ``CS + CS'' represents the class-specific attention structure (with $C$ queries) combined with class-agnostic classifier (with $C$ FCs).
It is interesting that the performance of ``CS+CA'' (68.7) is much lower than that of ``CA+CA'' (74.0), which seems like a more naive baseline than ``CS+CA''. When using the class-specific attention structure to obtain feature with shape of $200\times1280$, the FC classifier ($200\times1280$) must have a one-to-one correspondence with each class, \emph{i.e., } ``CS+CS'' (74.7), to achieve good results. If using one $1\times1280$ FC, \emph{i.e., } ``CS+CA'', to process all classes with the same parameters, discrimination power are insufficient and accuracy will decrease dramatically, which will be even lower than ``CA+CA''.

\subsection{Ablation study of $\alpha$ and $\beta$}
First, we explore the appropriate values for $\alpha$ and $\beta$, \emph{i.e.,} the ratios of $\mathcal{L}_{t\shortrightarrow v}$ and $\mathcal{L}_{v\shortrightarrow t}$ in \tabref{table:alpha} and \tabref{table3:beta}, respectively. We first fix $\alpha =0 $ to find the best $\beta$. As shown in \tabref{table:alpha}, as $\beta$ increases, the performance of both TQM and TSQNet rises up to a maximum at $\beta=0.6$ and then falls down. The performance of VQM remains unchanged, which demonstrates $\mathcal{L}_{t\shortrightarrow v}$ mainly benefit TQM in interactions. Then we fix $\beta=0.6$ to explore the impacts of $\alpha$.  As presented in \tabref{table3:beta}, the performance shows similar trend and the best results of TQM, VQM and TSQNet are achieved when $\alpha$ and $\beta$ both equal to 0.6, which implies $\mathcal{L}_{v\shortrightarrow t}$  benefits both TQM and VQM in interactions. After $\beta=0.6$, the performance of VQM breaks down, for prohibitively large $\beta$ hinders the convergence of the VQM. 

\begin{table}
\begin{minipage}[t]{0.47\linewidth}
\begin{center}
\caption{Ablation study of $\beta$ when fixing $\alpha=0$.}
\label{table:alpha}
\setlength{\tabcolsep}{3.0pt}
\begin{tabular}{cccc}
\toprule
$\beta$ & TSQNet& TQM  & VQM\\
\midrule
0.0 &  74.9 & 72.0 &74.6 \\
0.2 &  74.9 & 72.3 & 74.6\\
0.4 &  75.0 & 72.5 & 74.6\\
0.6 &  \textbf{75.1} & \textbf{72.7} & 74.6\\
0.8 &  74.8    & 72.6 & 74.6 \\
1.0 &  74.7   & 72.6 & 74.6 \\
% 0.0 0.6 &  74.6 & 72.4 & 74.3\\
\bottomrule
\end{tabular}
\end{center}
\end{minipage}
% \begin{table}
\hfill
\begin{minipage}[t]{0.47\linewidth}
\begin{center}
\caption{Ablation study of $\alpha$ when fixing $\beta=0.6$.}\label{table3:beta}
\setlength{\tabcolsep}{3.0pt}
\renewcommand{\arraystretch}{1.18}
\begin{tabular}{cccc}
\toprule
$\alpha$ & TSQNet & TQM & VQM \\
\midrule
0.0 &  75.1 & 72.7 & 74.6 \\
0.2 &  75.0 & 72.8 & 74.6 \\
0.4 &  75.1  & 72.8 & 74.7\\
0.6 &  \textbf{75.3}  & \textbf{73.1} & \textbf{74.8}\\
0.8 &  71.2  & 72.5 & 67.5\\
\bottomrule
\end{tabular}
\end{center}
\end{minipage}
\end{table}

\subsection{Detailed Ablation study for transformer decoder structures}
We further ablate the structure of the standard transformer decoder, \emph{viz.,} self-attention, number of layers and heads.  Typical transformer decoder contains a self-attention layer on the top of query matrix and multiple cross-attention layers with multi-head structure. In TSQNet, we use a quite brief version of transformer decoder, containing a single-head cross-attention layer without self-attention layers, to realize TSQ layer. Next we discuss the effectiveness of this design.  

\noindent \textbf{Impact of Self-attention layer.}
On one hand, self-attention layer on queries make each TSQ embedding interact with each other, which may cause the class-specific information to mix with each other and deviates the class-specific nature of TSQ embeddings. On the other hand, self-attention layers bring in extra computation complexity of $O(C^{2})$ ,  where $C$ is the number of categories.
% is not consistent with our class-specific motivation. 
% We did ablation of with and without self-attention to verify this in \tabref{table:selfattention}. 
As shown in \tabref{table:selfattention}, adding self-attention layer presents lower performance, which demonstrates that modelling relations between TSQ embeddings of categories can not produce better saliency measuring results. 
% And the experiment result shows that self-attention has a bad effect on the final result and increases the amount of computation.

\begin{table}
\centering
\begin{minipage}[t]{0.45\textwidth}
\centering
\caption{Ablation study of the usage of self-attention.}
\label{table:selfattention}
\renewcommand{\arraystretch}{1.3}
\begin{tabular}{cc}
\toprule
Methods & mAP (\%) \\
\midrule
w/ self-atten &  74.5  \\
w/o self-atten & \textbf{74.7} \\
\bottomrule
\end{tabular}
\end{minipage}
\hfill
\begin{minipage}[t]{0.45\textwidth}
\centering
\caption{Ablation study of Transformer Decoder layers and heads.}
\label{table:layerhead}
\begin{tabular}{cc}
\toprule
Methods & mAP (\%) \\
\midrule
1 layer 8 head & 73.7 \\
2 layer 1 head & 73.4 \\
1 layer 1 head &  \textbf{74.7}  \\
\bottomrule
\end{tabular}
\end{minipage}
\end{table}
% decoder的self attention

\noindent \textbf{Number of layers and heads.}
In TSQNet, the number of cross-attention layers and heads are both one. We present ablation experiments of more layers with more heads in \tabref{table:layerhead}. It is shown that both the increase of number of layers and heads make the mAP drop. For multiple cross-attention layers, the performance drop may attribute to lower discrepancy between queries in intermediate layers, which makes attention weights lack discrimination power between categories.
% The increase of number of layers which will brake the class-specific hypothesis of Q when do the cross attention at the last time, which lead to the decrease of mAP. 
For multi-head structure, the worse results may result from attention dimension splitting operation when calculating the similarity between query matrix and key matrix, which produces separate local similarities for multiple groups in feature dimension rather than the holistic similarity of the feature dimension.
% head 和 layer
% \subsection{Fusion analysis in other backbones and other datasets}
% To verify the impact of the fusion of VQM and TQM. Some fusion analysis in other backbones and other datasets is provide on \tabref{table:other_datset_fusion}.
% \begin{table}
% \begin{center}
% \caption{Fusion analysis in other backbones and other datasets.}
% \label{table:other_datset_fusion}
% \begin{tabular}{cccc}
% \toprule
% Dataset  & total mAP (\%) & text mAP (\%)  & vis mAP (\%)  \\
% \midrule
% ANet with Res50 & 76.5 & 74.6 & 75.8 \\
% FCVID & 83.5  & 82.0 & 83.1\\
% minik & & & \\
% \bottomrule
% \end{tabular}
% \end{center}
% \end{table}

\section{Additional Qualitative Analysis}\label{appendix:qualitive_analysis}
\begin{figure}[t] %H为当前位置，!htb为忽略美学标准，htbp为浮动图形
\centering %图片居中
\includegraphics[width=1.0\textwidth]{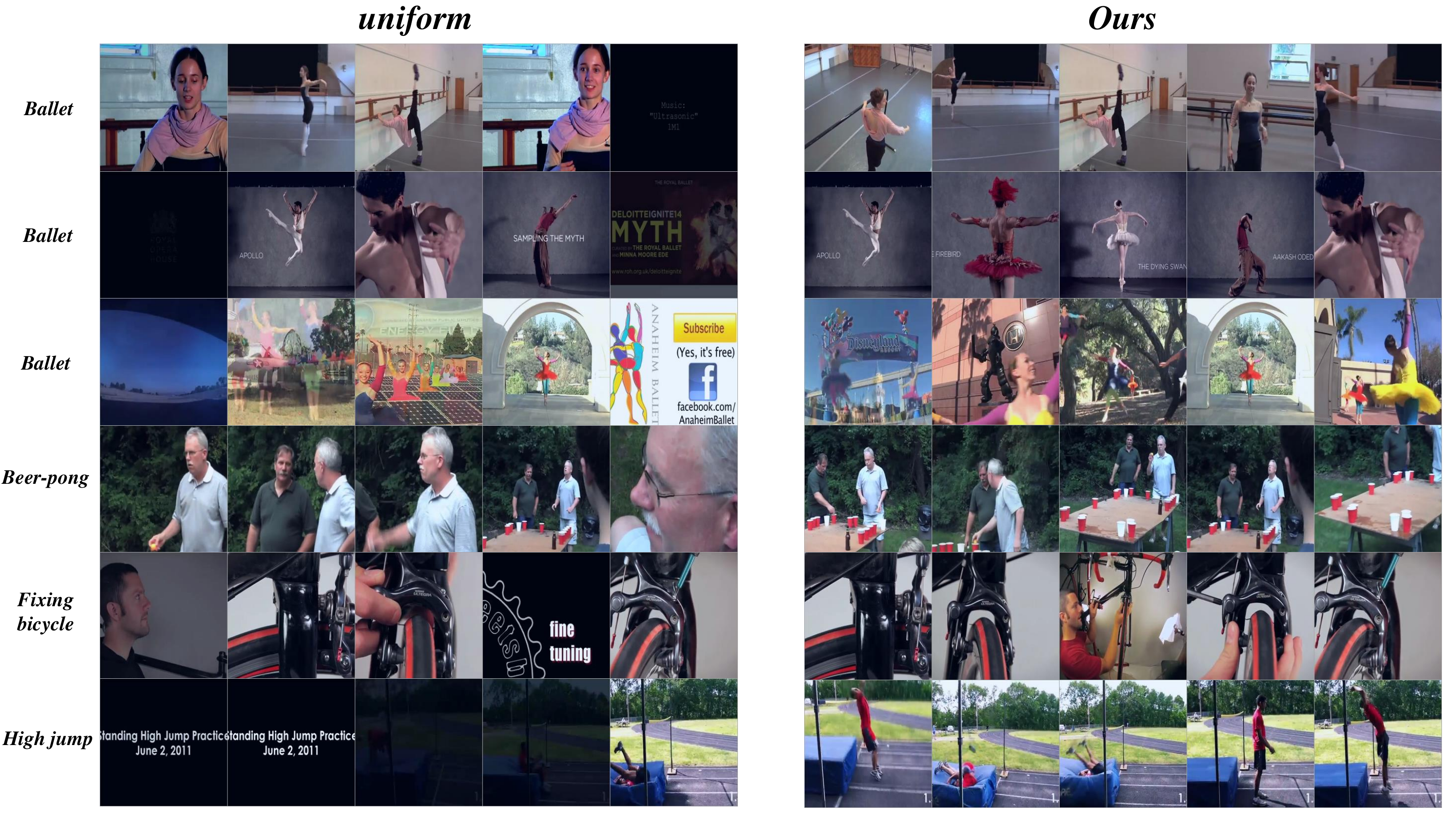} %插入图片，[]中设置图片大小，{}中是图片文件名
\caption{Qualitative Analysis on ActivityNet.} %最终文档中希望显示的图片标题
\label{Fig.anet} %用于文内引用的标签
\end{figure}

\begin{figure}[t] %H为当前位置，!htb为忽略美学标准，htbp为浮动图形
\centering %图片居中
\includegraphics[width=1.0\textwidth]{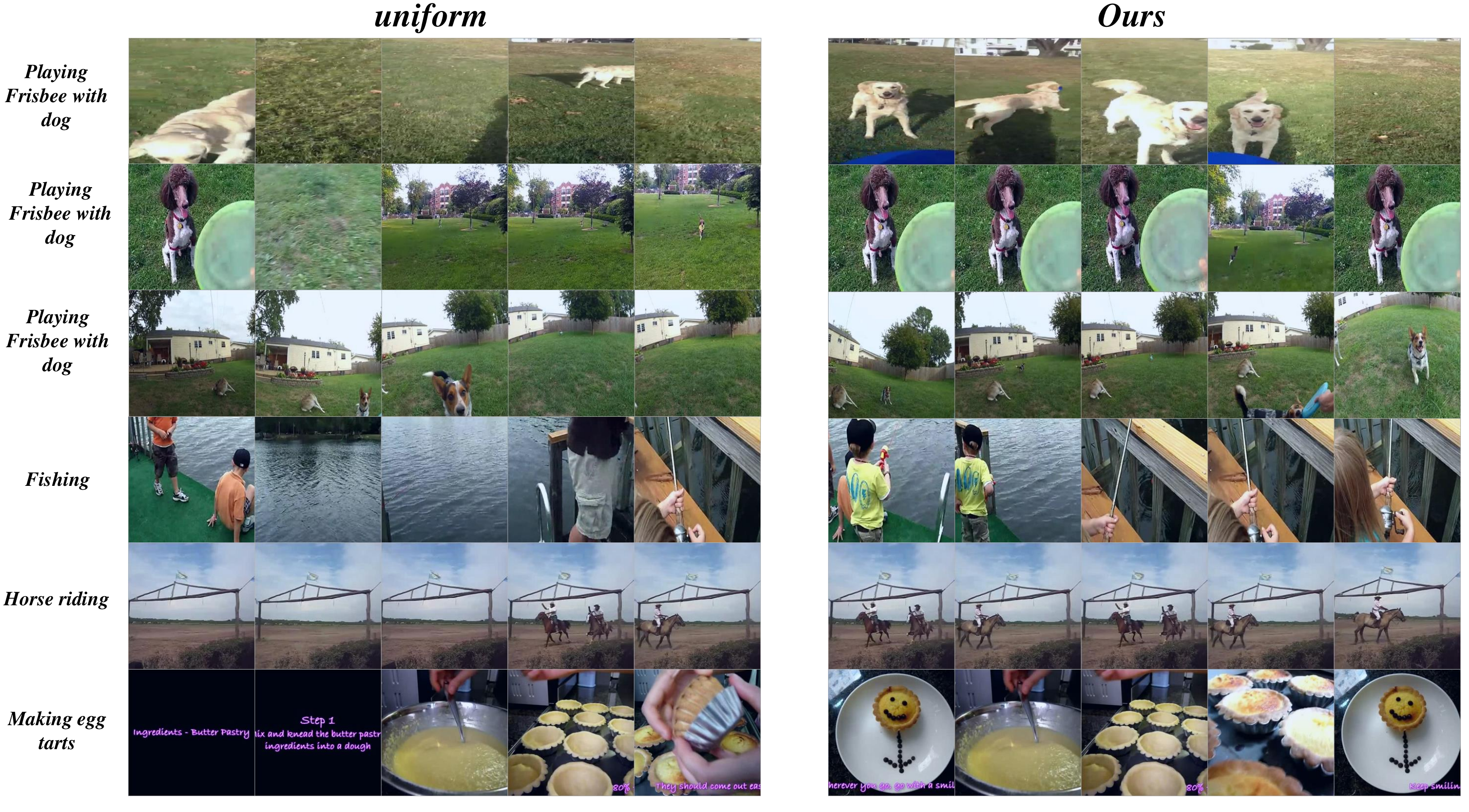} %插入图片，[]中设置图片大小，{}中是图片文件名
\caption{Qualitative Analysis on FCVID.} %最终文档中希望显示的图片标题
\label{Fig.fcvid} %用于文内引用的标签
\end{figure}

\figref{Fig.anet} and \figref{Fig.fcvid} show more qualitative results of TSQNet on ActivityNet and FCVID. For each dataset, we selected six examples, first three of which belongs to the same category and the last three  belongs to different categories. In \figref{Fig.anet}, we can see that our approach samples significantly more salient frames than the uniform baseline. 
% Uniform selects a lot of irrelevant frames, our approach focuses more on the event or action itself. 
Similar in \figref{Fig.fcvid}, uniform baseline selects many irrelevant frames, whereas our method selects more theme-related frames.

\section{Additional Visualization of TSQ Embeddings}
In this section, we provide the complete t-SNE visualization for TSQ embeddings of  both the VQM and TQM on ActivityNet to supplement the local zooms visualization in Section 4.5 of the main paper. Specifically, we visualize the start and end states of training for the two modules in two different initialization fashions, \emph{i.e.,} random and proposed initialization, respectively. For VQM, we compare the random initialization with the visual common appearance feature initialization. For TQM, we compare the random initialization with the class name Bert embedding feature initialization.

\begin{figure}[ht] %H为当前位置，!htb为忽略美学标准，htbp为浮动图形
\centering %图片居中
\includegraphics[width=0.95\textwidth]{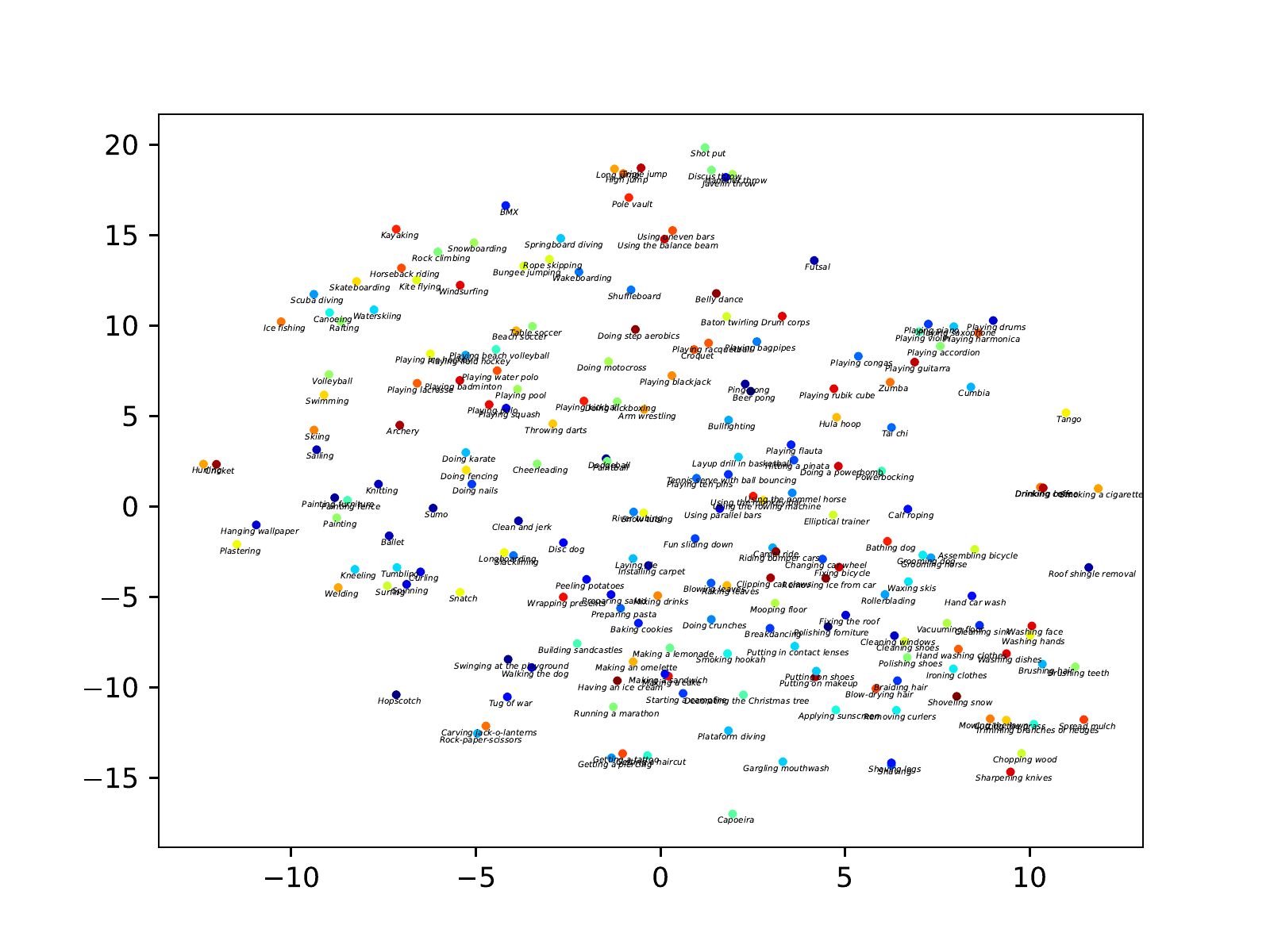} %插入图片，[]中设置图片大小，{}中是图片文件名
\caption{TSQ embeddings of TQM with class name Bert initialization before training.} %最终文档中希望显示的图片标题
\label{Fig.bert_start} %用于文内引用的标签
\end{figure}

\begin{figure}[t] %H为当前位置，!htb为忽略美学标准，htbp为浮动图形
\centering %图片居中
\includegraphics[width=0.95\textwidth]{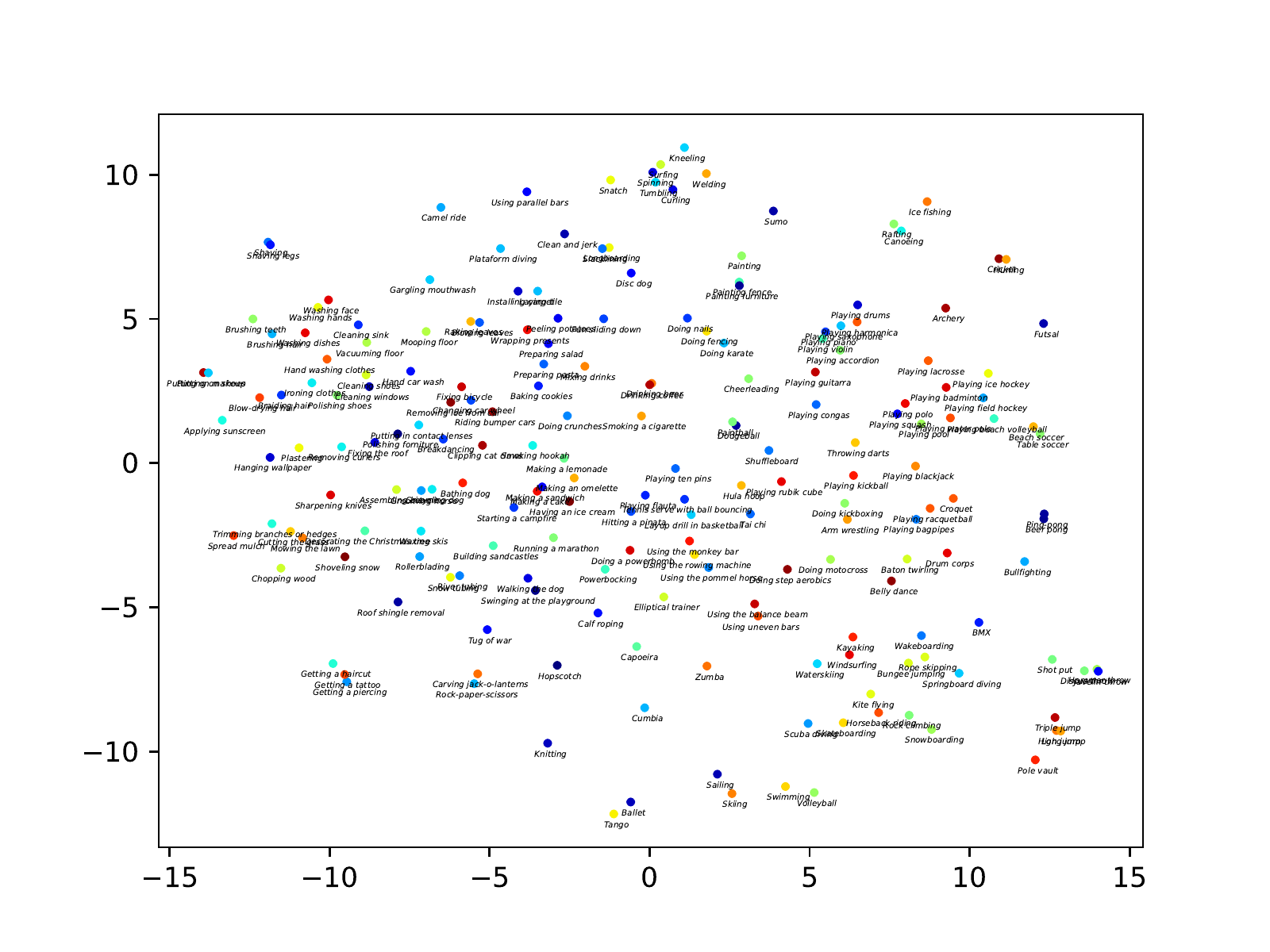} %插入图片，[]中设置图片大小，{}中是图片文件名
\caption{TSQ embeddings of TQM with class name Bert initialization after training.} %最终文档中希望显示的图片标题
\label{Fig.bert_best} %用于文内引用的标签
\end{figure}

\begin{figure}[t] %H为当前位置，!htb为忽略美学标准，htbp为浮动图形
\centering %图片居中
\includegraphics[width=0.95\textwidth]{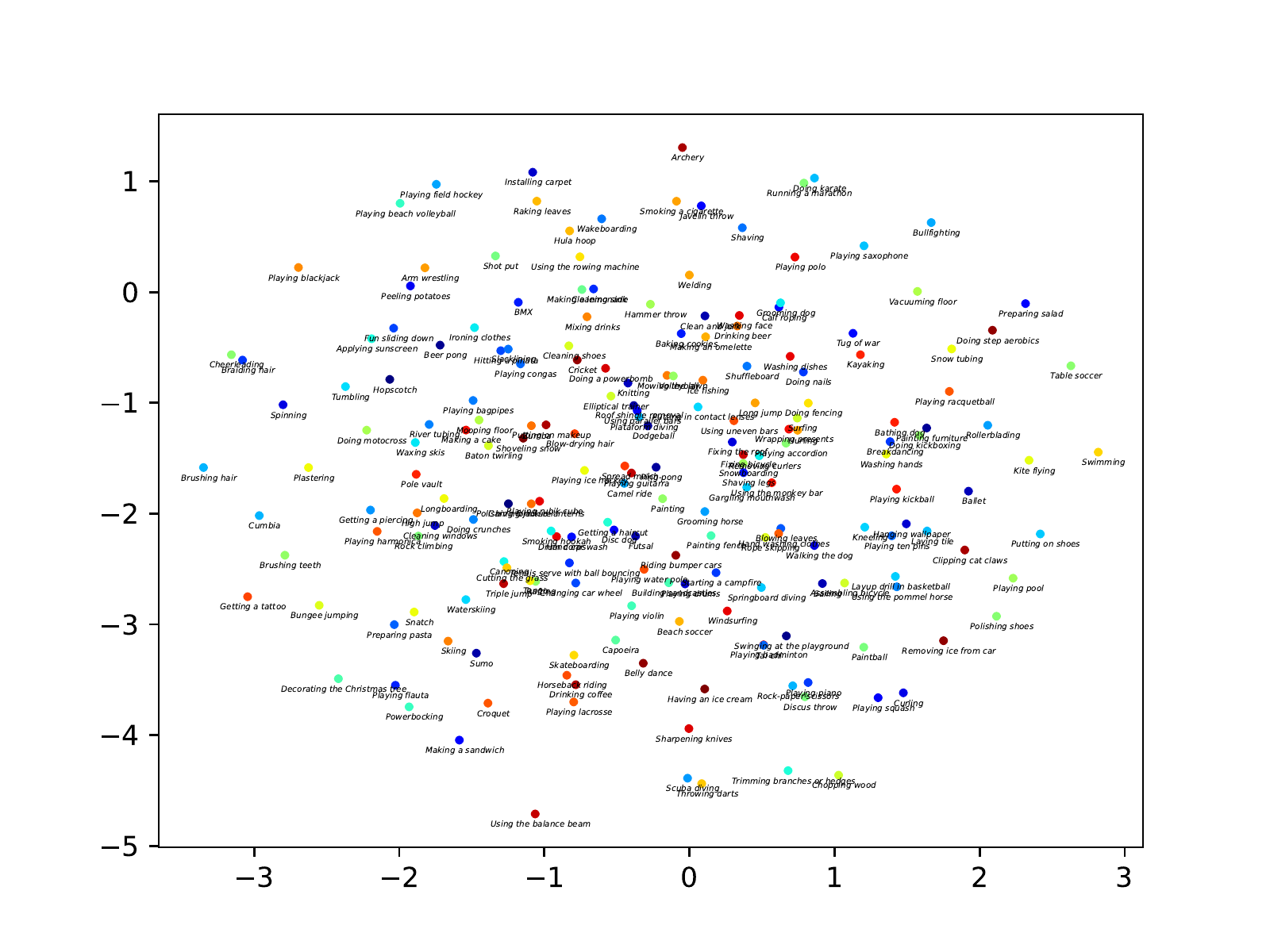} %插入图片，[]中设置图片大小，{}中是图片文件名
\caption{TSQ embeddings of TQM with random  initialization before training.} %最终文档中希望显示的图片标题
\label{Fig.random_start} %用于文内引用的标签
\end{figure}

\begin{figure}[t] %H为当前位置，!htb为忽略美学标准，htbp为浮动图形
\centering %图片居中
\includegraphics[width=0.95\textwidth]{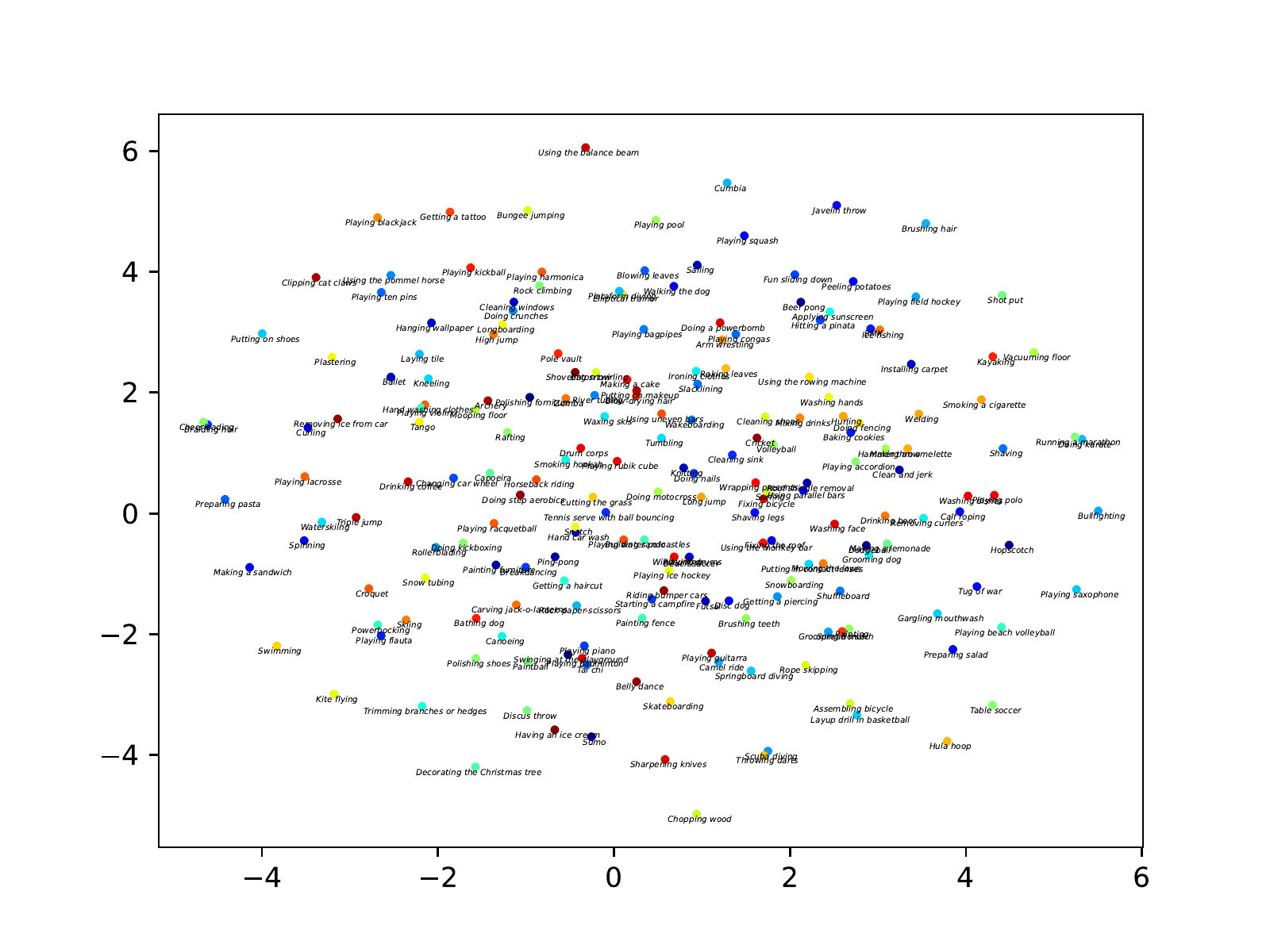} %插入图片，[]中设置图片大小，{}中是图片文件名
\caption{TSQ embeddings of TQM with random  initialization after training.} %最终文档中希望显示的图片标题
\label{Fig.random_best} %用于文内引用的标签
\end{figure}

\noindent \textbf{TSQ embeddings of TQM.}
\figref{Fig.bert_start} shows the visualization of TSQ embeddings of TQM with class name Bert embedding feature initialization \textbf{before} training.
\figref{Fig.bert_best} shows the visualization of TSQ embeddings of TQM with class name Bert embedding feature initialization \textbf{after} training.
\figref{Fig.random_start} shows the visualization of TSQ embeddings of TQM with random initialization \textbf{before} training.
\figref{Fig.random_best} shows the visualization of TSQ embeddings of TQM with random initialization \textbf{after} training.

\begin{figure}[t] %H为当前位置，!htb为忽略美学标准，htbp为浮动图形
\centering %图片居中
\includegraphics[width=0.92\textwidth]{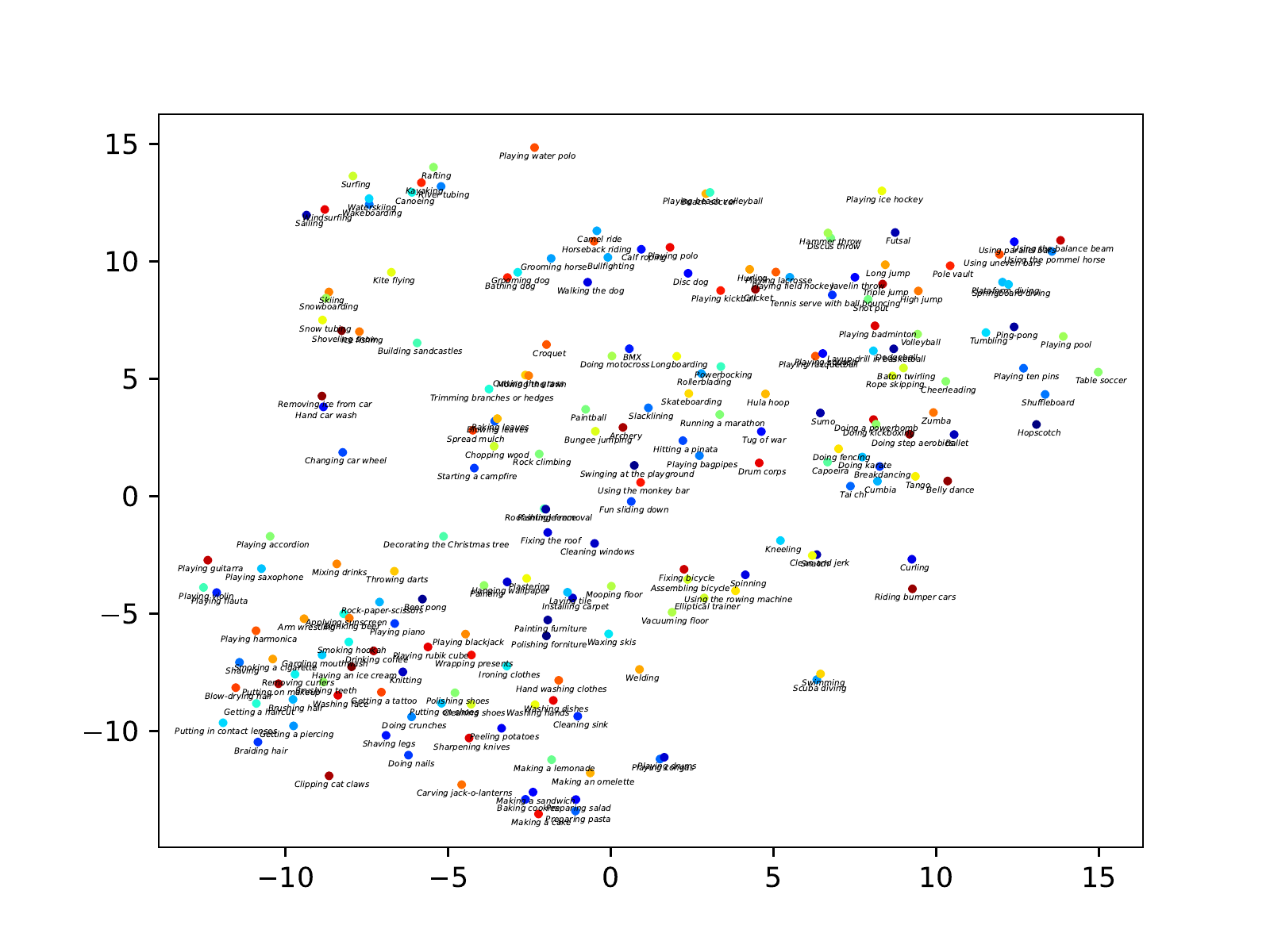} %插入图片，[]中设置图片大小，{}中是图片文件名
\caption{TSQ embeddings of VQM with appearance  initialization before training.} %最终文档中希望显示的图片标题
\label{Fig.vis_center_start} %用于文内引用的标签
\end{figure}

\begin{figure}[t] %H为当前位置，!htb为忽略美学标准，htbp为浮动图形
\centering %图片居中
\includegraphics[width=0.92\textwidth]{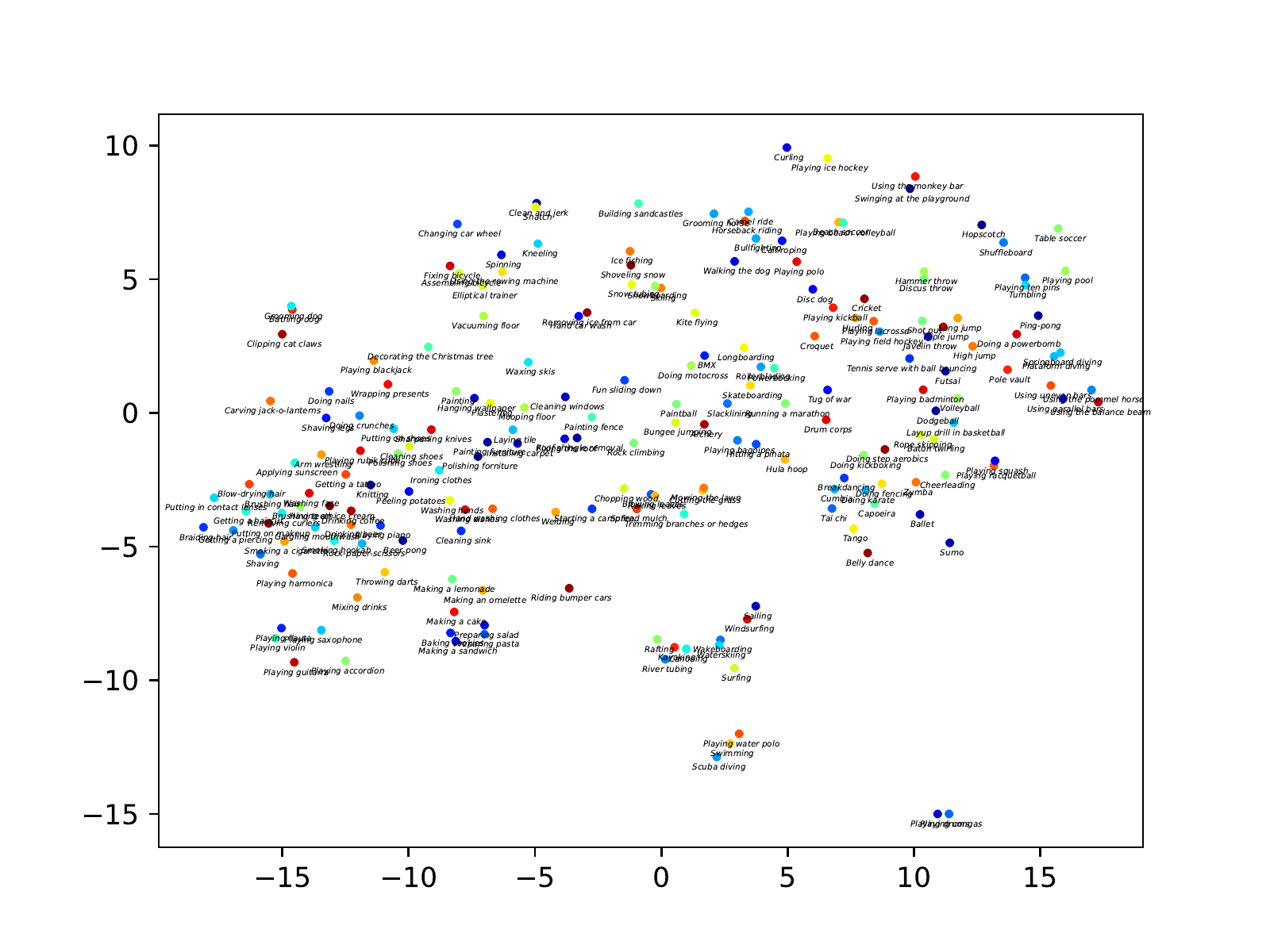} %插入图片，[]中设置图片大小，{}中是图片文件名
\caption{TSQ embeddings of VQM with common appearance  initialization after training.} %最终文档中希望显示的图片标题
\label{Fig.vis_center_best} %用于文内引用的标签
\end{figure}

\begin{figure}[t] %H为当前位置，!htb为忽略美学标准，htbp为浮动图形
\centering %图片居中
\includegraphics[width=0.95\textwidth]{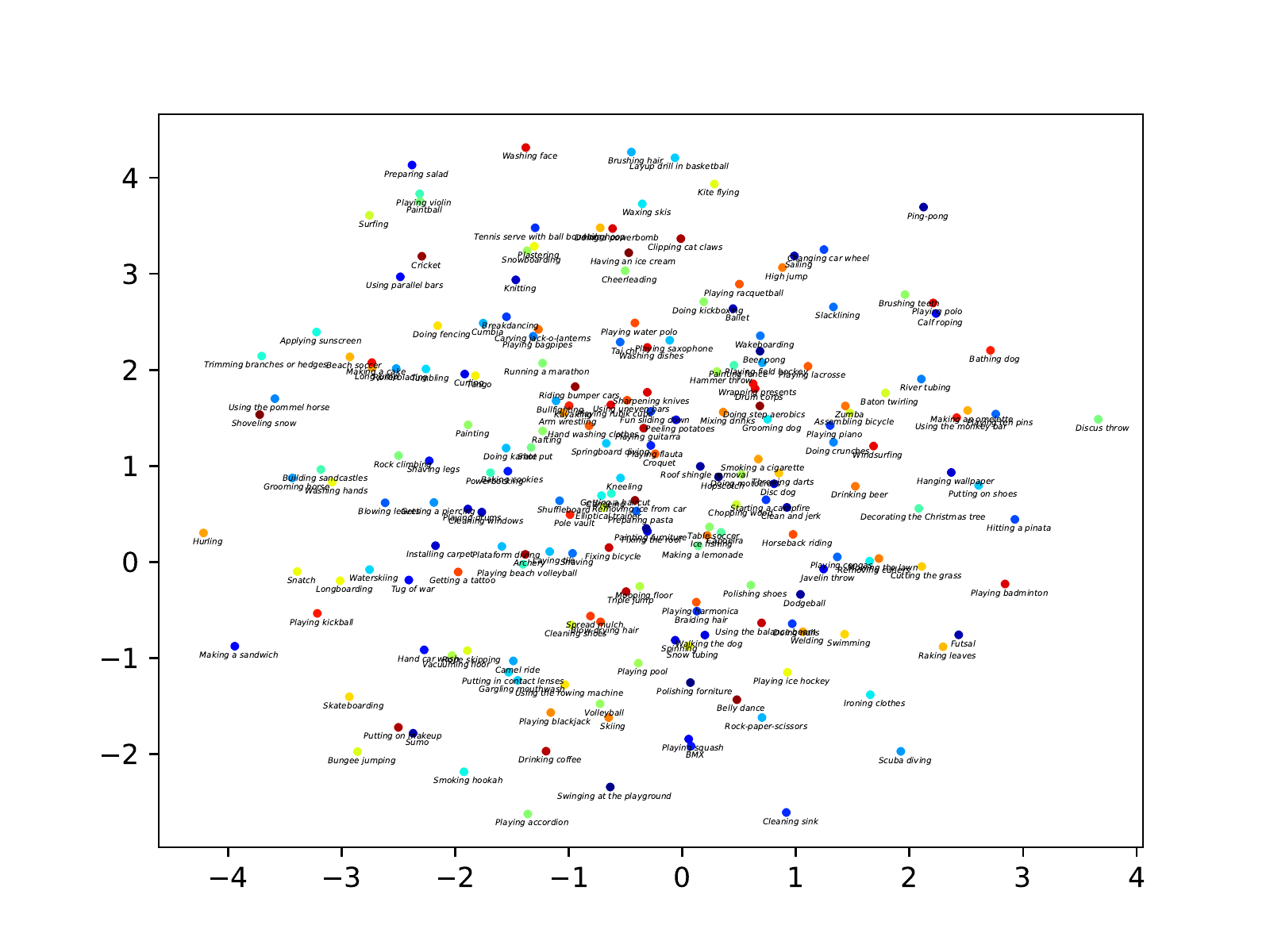} %插入图片，[]中设置图片大小，{}中是图片文件名
\caption{TSQ embeddings of VQM with random  initialization before training.} %最终文档中希望显示的图片标题
\label{Fig.vis_random_start} %用于文内引用的标签
\end{figure}

\begin{figure}[t] %H为当前位置，!htb为忽略美学标准，htbp为浮动图形
\centering %图片居中
\includegraphics[width=0.95\textwidth]{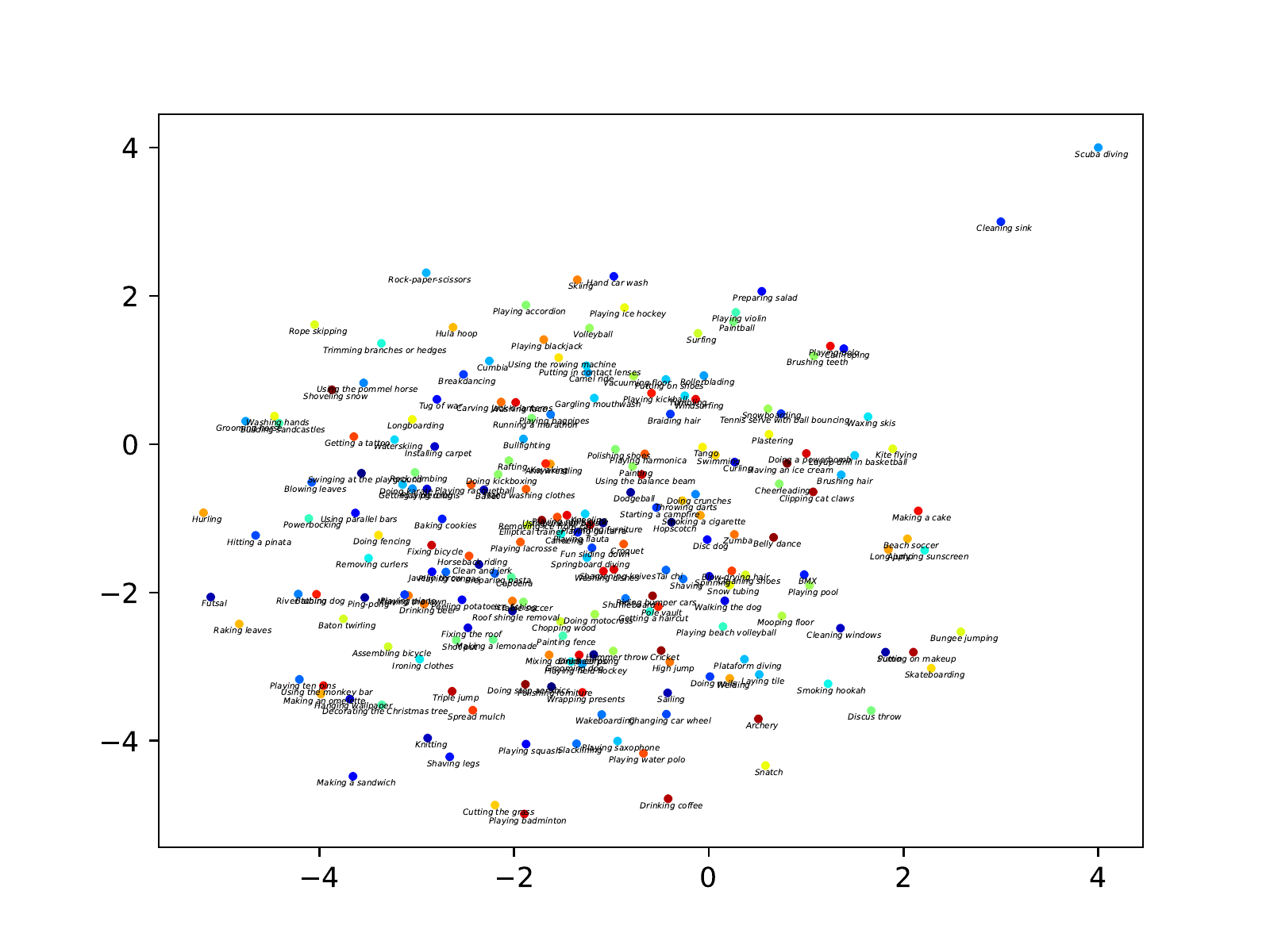} %插入图片，[]中设置图片大小，{}中是图片文件名
\caption{TSQ embeddings of VQM with random  initialization after training.} %最终文档中希望显示的图片标题
\label{Fig.vis_random_best} %用于文内引用的标签
\end{figure}

\noindent \textbf{TSQ embeddings of VQM.}
\figref{Fig.vis_center_start} shows the visualization of TSQ embeddings of VQM with common appearance feature initialization \textbf{before} training.
\figref{Fig.vis_center_best} shows the visualization of TSQ embeddings of VQM with common appearance feature initialization \textbf{after} training.
\figref{Fig.vis_random_start} shows the visualization of TSQ embeddings of VQM with random initialization \textbf{before} training.
\figref{Fig.vis_random_best} shows the visualization of TSQ embeddings of VQM with random initialization \textbf{after} training.

\clearpage
% ---- Bibliography ----
%
% BibTeX users should specify bibliography style 'splncs04'.
% References will then be sorted and formatted in the correct style.
%
\bibliographystyle{splncs04}
\bibliography{egbib}
\end{document}